\documentclass{article}

\usepackage[ruled,vlined]{algorithm2e}
\usepackage{microtype}
\usepackage{graphicx}
\usepackage{subfigure}
\usepackage{booktabs} % for professional tables

\usepackage{hyperref}

\usepackage[accepted]{icml2025}

\usepackage{amsmath}
\usepackage{amssymb}
\usepackage{mathtools}
\usepackage{amsthm}

\usepackage[T1]{fontenc}
\usepackage{pifont}
\usepackage{multirow}
\usepackage{makecell}
\usepackage{hyperref}
\usepackage{url}
\usepackage{array}
\usepackage{graphicx}
\usepackage{booktabs}
\usepackage{multirow}
\usepackage[most]{tcolorbox}
\usepackage{enumitem}
\usepackage[T1]{fontenc}
\usepackage{inconsolata}
\usepackage{wrapfig}
\usepackage{listings}
\usepackage{xcolor}

\newcommand{\fullname}{In-Context Preference Learning}
\newcommand{\name}{ICPL}

\newcommand{\rebuttal}[1]{{#1}}

\usepackage[capitalize,noabbrev]{cleveref}

\theoremstyle{plain}

\theoremstyle{definition}

\theoremstyle{remark}

\usepackage[textsize=tiny]{todonotes}

\icmltitlerunning{ICPL: Few-shot In-context Preference Learning via LLMs}

\begin{document}

\twocolumn[
% \icmltitle{Submission and Formatting Instructions for \\
           % International Conference on Machine Learning (ICML 2025)}
\icmltitle{\rebuttal{ICPL: Few-shot In-context Preference Learning via LLMs}}

% It is OKAY to include author information, even for blind
% submissions: the style file will automatically remove it for you
% unless you've provided the [accepted] option to the icml2025
% package.

% List of affiliations: The first argument should be a (short)
% identifier you will use later to specify author affiliations
% Academic affiliations should list Department, University, City, Region, Country
% Industry affiliations should list Company, City, Region, Country

% You can specify symbols, otherwise they are numbered in order.
% Ideally, you should not use this facility. Affiliations will be numbered
% in order of appearance and this is the preferred way.

\begin{icmlauthorlist}
\icmlauthor{Chao Yu}{yyy}
\icmlauthor{Qixin Tan}{yyy}
\icmlauthor{Hong Lu}{yyy}
\icmlauthor{Jiaxuan Gao}{yyy}
\icmlauthor{Xinting Yang}{yyy}
\icmlauthor{Yu Wang}{yyy}
\icmlauthor{Yi Wu}{yyy,comp}
%\icmlauthor{}{sch}
\icmlauthor{Eugene Vinitsky}{sch}
%\icmlauthor{}{sch}
%\icmlauthor{}{sch}
\end{icmlauthorlist}
        
\icmlaffiliation{yyy}{Tsinghua University, Beijing, China}
\icmlaffiliation{comp}{Shanghai Qi Zhi Institute, Shanghai, China}
\icmlaffiliation{sch}{New York University, New York, China}

\icmlcorrespondingauthor{Eugene Vinitsky}{vinitsky.eugene@gmail.com}

% You may provide any keywords that you
% find helpful for describing your paper; these are used to populate
% the "keywords" metadata in the PDF but will not be shown in the document
\icmlkeywords{Machine Learning, ICML}

\vskip 0.3in
]

% this must go after the closing bracket ] following \twocolumn[ ...

% This command actually creates the footnote in the first column
% listing the affiliations and the copyright notice.
% The command takes one argument, which is text to display at the start of the footnote.
% The \icmlEqualContribution command is standard text for equal contribution.
% Remove it (just {}) if you do not need this facility.

%\printAffiliationsAndNotice{}  % leave blank if no need to mention equal contribution
\printAffiliationsAndNotice{} % otherwise use the standard text.

\begin{abstract}
Preference-based reinforcement learning is an effective way to handle tasks where rewards are hard to specify but can be exceedingly inefficient as preference learning is often tabula rasa. 
% We demonstrate that Large Language Models (LLMs) have native preference-learning capabilities that allow them to achieve sample-efficient preference learning, addressing this challenge. 
\rebuttal{To address this challenge,
we propose In-Context Preference Learning (ICPL), which uses in-context learning capabilities of LLMs to reduce human query inefficiency.}
ICPL uses the task description and basic environment code to create sets of reward functions which are iteratively refined by placing human feedback over videos of the resultant policies into the context of an LLM and then requesting better rewards.
We first demonstrate ICPL’s effectiveness through a synthetic preference study, providing quantitative evidence that it significantly outperforms baseline preference-based methods with much higher performance and orders of magnitude greater efficiency. We observe that these improvements are not solely coming from LLM grounding in the task but that the quality of the rewards improves over time, indicating preference learning capabilities.
Additionally, we perform a series of real human preference-learning trials and observe that ICPL extends beyond synthetic settings and can work effectively with humans-in-the-loop. 
\end{abstract}
% ICPL takes the task description and basic environment code, synthesizes a set of reward functions, and 
% then repeatedly updates the reward functions using human feedback over videos of the resultant policies over a small number of trials. 
% to accelerate learning reward functions from preferences. 
% We investigate whether  can  We propose 

\section{Introduction}\label{sec:intro}
% Desired agent behaviors are often hard to define explicitly: How should one code a reward for "follow instructions" or "behave in a non-toxic manner"? Preference-based learning, in which rewards are learned by comparing orderings over trajectories, has emerged as a solution to this challenge, as it allows us to sidestep the challenge of reward specification.
% While it is hard to imagine how one could explicitly specify behaviors like "follow instructions", it is easy to label whether an instruction has or has not been obeyed. 
% The efficacy of preference learning has made it an essential component of many deployed systems, being used for XYZ.
% However, despite its appealing properties, preference learning has been challenging to deploy and scale as collecting data for preference learning is a clear bottleneck. 
% Even simple tasks when learned tabula rasa, such as learning a reward corresponding to pressing a button, can require over $10000$ labels~\citep{lee1b} to achieve good performance.
% While directly encoding behaviors like “following instructions” is difficult, determining whether an instruction has been followed is much more straightforward.
Defining desired agent behaviors explicitly is often challenging—how does one design a reward function for “following instructions” or “behaving in a non-toxic manner”? Preference-based reinforcement learning (PbRL) offers a solution by learning rewards through comparisons of trajectory orderings, bypassing the need for explicit reward specification. While directly encoding behaviors like “following instructions” is difficult, determining whether an instruction has been followed is much more straightforward. 
PbRL, in which we simultaneously learn a policy and reward that satisfy a set of preferences, has shown success in various tasks~\citep{christiano2017deep,ibarz2018reward,liu2020learning,wu2021recursively,lee2021pebble}. However, its deployment and scalability remain limited by the costly data collection process, which serves as a major bottleneck. Even for simple tasks, such as learning a reward for pressing a button, it requires over 10k labeled comparisons~\citep{lee1b} to achieve good performance as PbRL is often tabula rasa.

We investigate whether in-context learning (ICL), the ability of LLMs to modify their behavior based on a few provided examples, can be used to make the preference-learning step of PbRL sample efficient. Specifically, we focus on \emph{online} PbRL, where data collection and reward function training occur in an iterative loop. In each round, we collect preference data, use it to learn a reward function, learn a policy under that reward, and then gather new preference data under the updated policy. 
% In particular, we focus on \emph{online} preference learning (hereafter just referred to as preference learning) in which we alternate between collecting data and training reward functions. In each round, we collect data, use it to learn a reward function, and then collect new data under the new reward function. This process is repeated iteratively with the goal of continually improving the quality of the learned reward. 

To incorporate ICL into the PbRL loop, we put into context the collected series of preferences and programmatic reward functions and ask the LLM to compare across the context to infer a new reward that better explains the preferences. Asking this of an LLM goes beyond previously observed ICL capabilities, requiring not just imitation of provided examples that have been stored in context but also reasoning about the relationships between different rewards and their alignment with preferences. Moreover, determining the optimal way to structure this information to maximize preference learning performance remains an open question.

As a step towards LLMs functioning as effective guides of the PbRL loop, we introduce a new method, {\fullname} ({\name}), which significantly enhances the human query efficiency of PbRL. Our approach is to harness the coding capabilities of LLMs to autonomously generate improved reward functions, represented as code, that take into context prior preferences and generated rewards. Specifically,  {\name} leverages an LLM to generate executable, diverse reward functions based on the task description and environment source code. We acquire preferences by evaluating the agent behaviors resulting from these reward functions, selecting the most and least preferred behaviors. The selected functions, along with historical data such as reward traces of the generated reward functions from RL training, are then fed back into the LLM which attempts to generate an improved reward function. 
\rebuttal{Note that unlike other instances in which an LLM is used to generate rewards~\cite{ma2023eureka}, there is no ground-truth metric, such as sparse reward, that the LLM can use to evaluate agent performance, and thus, success here would demonstrate that preference learning is occurring.}
% LLMs have some native preference-learning capabilities. 

% We hypothesize that as a result of the in-context learning capabilities of LLMs and their grounding in text data that contains descriptions of tasks, {\name} will be able to improve the quality of the reward function through incorporating the preferences and the history of the generated reward functions, ensuring they align more and more closely with human preferences. 
% Unlike evolutionary search methods like EUREKA~\cite{ma2023eureka}, 
% \rebuttal{is this reasonable here?}

To study the effectiveness of {\name}, we perform experiments on a diverse set of RL tasks. For scalability of experimentation, we first study tasks with proxy human preferences where a ground-truth reward function is used to assign preference labels. We observe that compared to traditional PbRL algorithms, {\name} achieves over a 30 times reduction in the required number of preference queries to achieve equivalent or superior performance. Moreover, given only preference feedback, {\name} achieves performance comparable to reward-generation methods that require %utilize 
a ground-truth sparse reward as feedback~\citep{ma2023eureka}. Additionally, we perform a series of real human preference learning trials and observe that ICPL extends beyond synthetic settings and can work effectively with humans-in-the-loop. Finally, we test {\name} on a particularly challenging task, ``making a humanoid jump like a real human,'' where designing a reward is difficult. By using real human feedback, our method successfully trained an agent capable of bending both legs and performing stable, human-like jumps, showcasing the potential of {\name} in tasks where human intuition plays a critical role.

In summary, the contributions of the paper are the following:
\begin{itemize}
    \item We propose {\name}, an LLM-based online preference learning algorithm. Over a synthetic set of preferences, we demonstrate that {\name} can iteratively output rewards that increasingly reflect preferences. Via a set of ablations, we demonstrate that this improvement is relatively monotonic, suggesting that preference learning is occurring as opposed to a random search and that the tasks are not memorized.
    \item We demonstrate that {\name} sharply outperforms tabula-rasa PbRL methods in terms of query efficiency and performance and is also competitive with reward-generation methods that rely on access to a ground-truth sparse reward.
    \item Through human-in-the-loop trials, we demonstrate that {\name} performs effectively even when dealing with significantly noisier preference labels.
\end{itemize}

\vspace{-2mm}
\section{Related Work}\label{sec:related}
\vspace{-2mm}
% \textbf{Human-in-the-loop Reinforcement Learning.}
Feedback from humans has been proven to be effective in training RL agents that better match human preferences~\citep{retzlaff2024human,mosqueira2023human,kwon2023reward}. Previous works have investigated human feedback in various forms, such as trajectory comparisons, preferences, demonstrations, and corrections~\citep{wirth2017survey,ng2000algorithms,jeon2020reward,peng2024learning}.
Among these various methods, PbRL learns both a reward model and policy based on human preferences across different trajectories~\citep{liu2020learning,wu2021recursively} and has been successfully scaled to train large foundation models for hard tasks like dialogue, e.g. ChatGPT~\citep{ouyang2022training}. 
In LLM-based applications, prompting is a simple way to provide human feedback in order to align LLMs with human preferences~\citep{giray2023prompt,white2023prompt,chen2023unleashing}. 
Iteratively refining the prompts with feedback from the environment or human users has shown promise in improving the output of the LLMs~\citep{wu2021recursively,nasiriany2024pivot}.
This work builds on these demonstrations of the ability to control LLM behavior via in-context prompts. We use interactive rounds of preference feedback between the LLM and humans to guide the LLM to generate reward functions that gradually elicit behaviors that align with human preferences.

% Our exploration that LLMs can be used to generate novel reward functions through in-context learning build upon several prior demonstrated LLM capabilities. 
Our investigation into using LLMs to generate novel reward functions via in-context learning builds on several previously demonstrated capabilities of LLMs.
First, it has been previously observed that LLMs can translate textual descriptions of desired behaviors into reward functions, expressed as code, that approximate those behaviors~\citep{ma2022vip, du2023vision, karamcheti2023language, kwon2023reward, wang2024rl,ma2024dreureka,10661065}. This suggests some useful prior knowledge that is helpful in incorporating text descriptions of the task into the subsequent generated code.
% it has been previously observed that LLMs can go from text descriptions of a desired behavior to rewards, represented as code, that approximately implements that behavior. 
Second, prior works have demonstrated that LLMs can use a variety of signals such as a history of rewards to generate new programs that improve on the history of rewards~\cite{ma2023eureka,romera2024mathematical}. This suggests that LLMs can function as approximate optimizers, a capability that forms the foundation of our approach to preference learning.
% This demonstrated that LLMs can approximately serve as optimizers, a capability that underlies our usage of it for preference learning.
Finally, LLMs have been demonstrated to be able to perform complex functional transformations on data stored in their context, as is required for preference learning, being able to perform linear regression~\cite{tang2024understanding}, no-regret learning~\cite{park2024llm}, and combinatorial problems~\cite{10611913}.

In concurrent work~\cite{clark2025efficientlygeneratingexpressivequadruped}, LLMs generate task parameterizations for quadruped locomotion, with humans ranking trajectories to identify optimal configurations via preference learning. Both papers noted that LLMs can be used to perform in-context preference learning, their approach focuses on generating task parameterizations as reward function vectors for gradient-based optimization, while our ICPL framework shows that LLMs can directly act as few-shot preference learners to generate and optimize reward functions.

\section{Problem Definition}\label{sec:prelimary}
% \vspace{-3mm}
% Reward design problem with human preference function
\begin{figure*}[htbp]
    \centering
    % \vspace{-4mm}
    \includegraphics[width=0.85\linewidth]{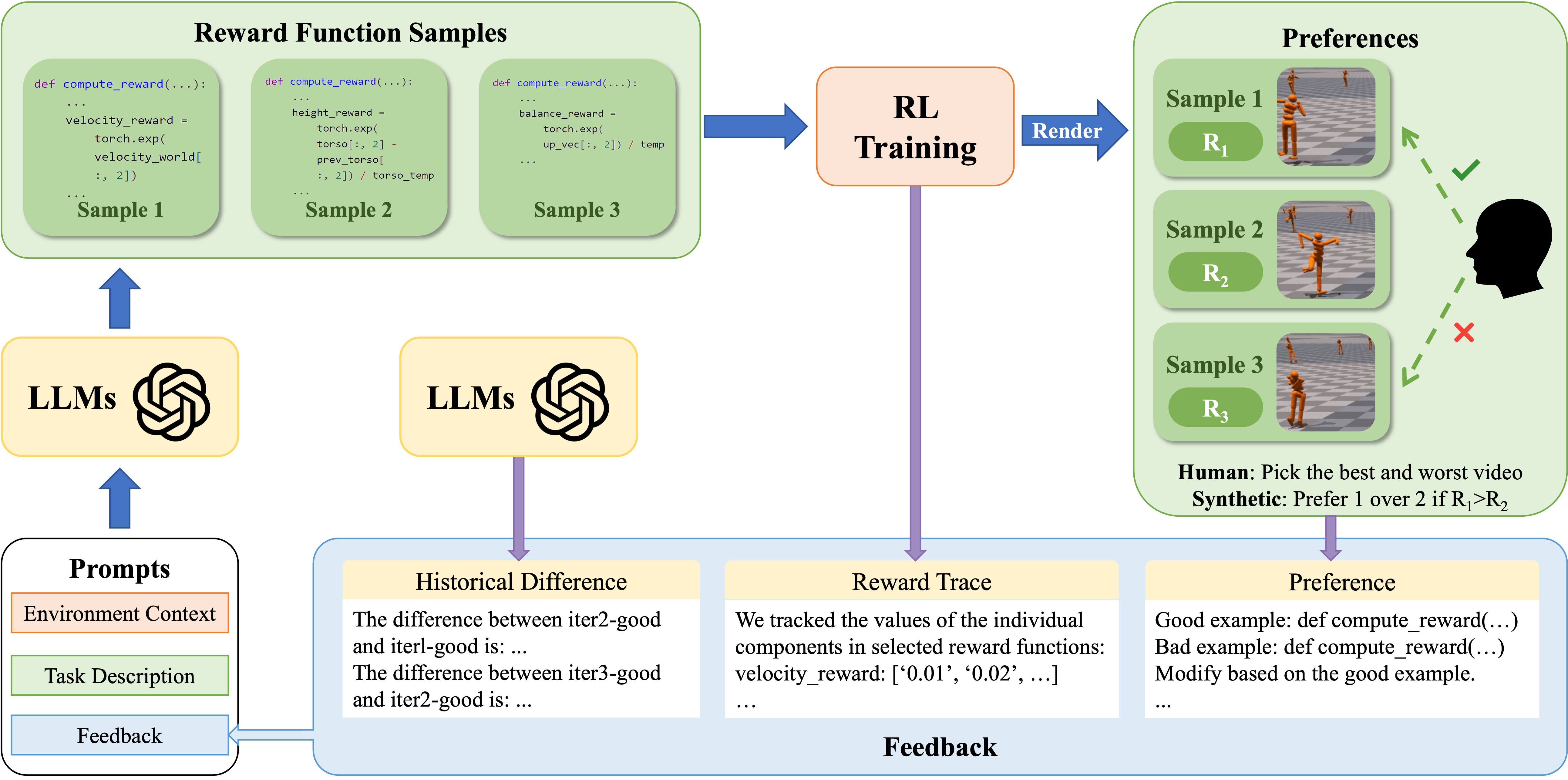}
    \caption{\name\ employs the LLM to generate initial $K$ executable reward functions based on the task description and environment context. Using RL, agents are trained with these reward functions. Videos are generated of the resultant agent behavior from which human evaluators select their most and least preferred. These selections serve as examples of positive and negative preferences. The preferences, along with additional contextual information, are provided as feedback prompts to the LLM, which is then requested to synthesize a new set of reward functions. For experiments simulating human evaluators, task scores are used to determine the best and worst reward functions. }
    % \vspace{-3mm}
    \label{fig: overview}
\end{figure*}
Our goal is to design a reward function that can be used to train RL agents that demonstrate human-preferred behaviors. It is usually hard to design proper reward functions in reinforcement learning that induce policies that align well with human preferences. 

\textbf{Markov Decision Process with Preferences~\cite{wirth2017survey}.} A \emph{Markov Decision Process with Preferences} (MDPP) is defined as a tuple $M=\langle\mathcal S, A, \mu,\sigma, \gamma,\rho\rangle$ where $\mathcal S$ denotes the state space, $A$ denotes the action space, $\mu$ is the distribution of initial states, $\sigma$ is the state transition model, $\gamma\in[0,1)$ is the discount factor. $\rho$ is the preference relation over trajectories, i.e. $\rho(\tau_i\succ\tau_j)$  denotes the probability with which trajectory $\tau_i$ is preferred over $\tau_j$. Given a set of preferences $\zeta$, the goal in an MDPP is to find a policy $\pi^*$ that maximally complies with $\zeta$. A preference $\tau_1\succ\tau_2$ is satisfied by $\pi$ if and only if $\text{Pr}_\pi(\tau_1)> \text{Pr}_\pi(\tau_2)$ where $\text{Pr}_\pi(\tau)=\mu(s_0)\prod_{t=0}^{|\tau|}\pi(a_t|s_t)\sigma(s_{t+1}|s_t,a_t)$. This can be viewed as finding a $\pi^*$ that minimizes a preference loss $L(\pi_\zeta)=\sum_iL(\pi,\zeta_i)$, where $L(\pi,\tau_1\succ\tau_2)=-(\text{Pr}_\pi(\tau_1)-\text{Pr}_{\pi}(\tau_2))$.

\textbf{Reward Design Problem with Preferences.} A \emph{reward design problem with preferences (RDPP)} is a tuple $P=\langle M,\mathcal R, A_M, \zeta\rangle$, where $M$ is a Markov Decision Process with Preferences, $\mathcal R$ is the space of reward functions, $A_M(\cdot):\mathcal R\rightarrow \Pi$ is a learning algorithm that outputs a policy $\pi$ that optimizes a reward $R\in\mathcal R$ in the MDPP. $\zeta=\{(\tau_1,\tau_2)\}$ is the set of preferences. In an RDPP, the goal is to find a reward function $R\in\mathcal R$ such that the policy $\pi=A_M(R)$ that optimizes $R$ maximally complies with the preference set $\zeta$. In PbRL, the learning algorithms usually involve multiple iterations, and the preference set $\zeta$ is constructed in every iteration by sampling trajectories from the policy or policy population.

% \vspace{-3mm}
\section{Method}\label{sec:method}
% \vspace{-3mm}

Our proposed method, In-Context Preference Learning (\name), integrates LLMs with human preferences to synthesize reward functions. The LLM receives environmental context and a task description to generate an initial set of $K$ executable reward functions. {\name} then iteratively refines these functions. In each iteration, the LLM-generated reward functions are trained within the environment, producing a set of agents; we use these agents to generate videos of their behavior. A ranking is formed over the videos, from which we retrieve the best and worst reward functions corresponding to the top and bottom videos in the ranking. These selections serve as examples of positive and negative preferences. The preferences, along with additional contextual information, such as reward traces and differences from previous good reward functions, are provided as feedback prompts to the LLM. The LLM takes in this context and is asked to generate a new set of rewards. 
% Algo.~\ref{algo: icpl} presents the pseudocode, and 
Fig.~\ref{fig: overview} illustrates the overall process of {\name}.
\subsection{Reward Function Initialization}

To enable the LLM to synthesize effective reward functions, it is essential to provide task-specific information, which consists of two key components: a description of the environment, including the observation and action space, and a description of the task objectives. At each iteration, {\name} ensures that $K$ executable reward functions are generated by resampling until there are $K$ executable reward functions.

\vspace{-2mm}
\subsection{Search Reward Functions by Human Preferences}
\vspace{-2mm}

For tasks without reward functions, the traditional PbRL typically involves constructing a reward model, which often demands substantial human feedback. 
Our approach, {\name}, aims to enhance efficiency by leveraging LLMs to directly search for optimal reward functions without the need to learn a reward model.
To expedite this search process, we use an LLM-guided search to find well-performing reward functions. Specifically, we generate $K=6$ executable reward functions per iteration across $N=5$ iterations. In each iteration, humans select the most preferred and least preferred videos, resulting in a good reward function and a bad one. These are used as a context for the LLM to use to synthesize a new set of $K$ reward functions. These reward functions are then used in a PPO~\citep{schulman2017proximalpolicyoptimizationalgorithms} training loop, and videos are rendered of the final trained agents.
% \vspace{-2mm}
\subsection{Automatic Feedback}
% \vspace{-2mm}
In each iteration, the LLM not only incorporates human preferences but also receives automatically synthesized feedback.
This feedback is composed of three elements: the evaluation of selected reward functions, the differences between historical good reward functions, and the reward trace of these historical reward functions.

\textbf{Evaluation of reward functions}: The component values that make up the good and bad reward functions are obtained from the environment during training and provided to the LLM. This helps the LLM assess the usefulness of different parts of the reward function by comparing the two.

\textbf{Differences between historical reward functions}: The best reward functions selected by humans from each iteration are taken out, and for any two consecutive good reward functions, their differences are analyzed by another LLM. These differences are supplied to the primary LLM to assist in adjusting the reward function.

\textbf{Reward trace of historical reward functions}: The reward trace, consisting of the values of the good reward functions during training from all prior iterations, is provided to the LLM. This reward trace enables the LLM to evaluate how well the agent is actually able to optimize those reward components.
%This reward trace enables the LLM to evaluate whether the modifications to the reward function have led to improvements.

%the information regarding the good and bad reward functions selected by humans from the environment, the differences between history reward functions, and the reward traces of these history reward functions.

%For the first part, we obtain the values of the components that make up the good and bad reward functions during training from the environment and provide them to the LLM. This helps the LLM to determine the usefulness of each part of the reward function and optimize it.

%For the second part, we take out the good reward functions in each iteration, and for any two good reward functions from the continuous iterations, we provide them to another LLM to get the difference between them. All these differences are provided to the LLM so that it can optimize with the reward function changes.

%For the third part, the values of the components of the good reward functions during training from all the previous iterations are provided to the LLM. Such a reward trace helps the LLM to check whether changes to the reward function have resulted in enhancements.

\vspace{-4mm}
\section{Experiments}\label{sec:exp}
In this section, we conducted two sets of experiments to evaluate the effectiveness of our method: one using proxy human preferences and the other using real human preferences. 

1) \textbf{Proxy Human Preference}:
We follow the standard experiment setting of PbRL where human-designed rewards were used as proxies of human preferences. 
% Specifically, if ground truth reward $R_1 > R_2$, sample 1 is preferred over sample 2. 
It enables rapid and quantitative evaluation of our approach. Proxy human preference corresponds to a noise-free case that is likely easier than human trials; if {\name} performed poorly here it would be unlikely to work in human trials. 
% The human-designed rewards were employed to automatically classify good and bad samples, allowing for rapid, quantitative evaluations of our method. 
% This approach was chosen because real human experiments are time-consuming and not scalable.
Importantly, human-designed rewards were only used to automate the selection of samples and were not included in the prompts sent to the LLM; the LLM \textbf{never observes the functional form of the ground truth rewards nor does it ever receive any values from them}. Since proxy human preferences are free from noise, they offer a reliable comparison to evaluate our approach efficiently. However, as discussed later in the limitations section, these proxies may not correctly measure challenges in human feedback such as the inability to rank samples, intransitive preferences, or other biases.

2) \textbf{Human-in-the-loop Preference}:
To further validate our method, we conducted a second set of experiments with human participants. These participants repeated the tasks from the Proxy Human Preferences and engaged in an additional task that lacked a clear reward function.
% : ``Making a humanoid jump like a real human.''

\subsection{Baselines}
We consider three PbRL methods as baselines, which update reward models during training. B-Pref~\citep{lee1b}, a benchmark specifically designed for PbRL, provides two of our baseline algorithms: \textbf{PrefPPO} and \textbf{PEBBLE}. PrefPPO is based on the on-policy RL algorithm PPO, while PEBBLE builds upon the off-policy RL algorithm SAC.
Additionally, we include \textbf{SURF}~\citep{park2022surf}, which enhances PEBBLE by utilizing unlabeled samples with data augmentation to improve feedback efficiency. For each task, we use the default hyperparameters of PPO and SAC provided by IsaacGym, which were fine-tuned for high performance. This ensures a fair comparison across methods. Further details can be found in Appendix \ref{app: baseline}.

\textbf{Definition of Human Query $Q$.} To evaluate the human effort required for {\name} and baseline methods, we track the number of human queries $Q$, which quantifies the amount of human effort involved in a human-in-the-loop experiment—a crucial factor for these methods. Specifically, we define a single human query as a human comparing two trajectories or videos and providing a preference.

In {\name}, each iteration generates $K$ reward function samples, resulting in $K$ corresponding videos. The human compares these videos, first selecting the best one, then picking the worst from the remaining $K-1$ videos. After $N=5$ iterations, the best video of each iteration is compared to select the overall best. The number of human queries $Q$ can be calculated as $Q = (K-1) \times 2N - 1 $. For {\name}, with $K=6$ and $N=5$, this results in $Q=49$. 
In baseline methods, humans compare two sampled trajectories and provide a preference label to update the reward model. To ensure a fair comparison, we set the maximum number of queries to  $Q=49$, matching {\name}. Additionally, we evaluate larger query budgets of $Q=150, 1.5k, 15k$, denoted as Baseline-\#$Q$.
%, to assess their impact on the final task score (FTS) across different tasks.

% \vspace{-2mm}
\subsection{Testbed}
\textbf{Tasks.} We first adopt several tasks from the GPU-based IsaacGym with human-designed rewards for quantitative comparison~\citep{ma2023eureka}, covering diverse environments: \textit{Cartpole, BallBalance, Quadcopter, Anymal, Humanoid, Ant, FrankaCabinet, ShadowHand,} and \textit{AllegroHand}. To ensure fair evaluation, we strictly follow the original task configurations, including observation space, action space, and reward computation. We refer to these tasks collectively as \textit{IsaacGym Tasks} in the following discussion.
Additionally, we introduce a new task, HumanoidJump, defined as ``making a humanoid jump like a real human.'' Defining a clear reward for this task is challenging, as human-like jumping lacks easily quantifiable criteria.

\textbf{Task Metric.} Here, we provide a specific explanation of how sparse rewards (detailed in Appendix \ref{app: env}) are used as task metrics in the adopted IsaacGym tasks. 
The task metric is the average of the sparse rewards across parallel environments. 
To assess the generated reward function or the learned reward model for each RL run, we take the maximum task metric value sampled at fixed intervals, marked as \textit{task score of reward function/model} (RTS). In each iteration, {\name} generates 6 RL runs and selects the highest RTS as the result for that iteration. {\name} performs 5 iterations and then selects the highest RTS from these iterations as the \textit{task score} (TS) for each experiment. Due to the inherent randomness of LLMs, we run 5 experiments for all methods, and report the highest TS as the \textit{final task score} (FTS) for each approach. A higher FTS indicates better performance across all tasks.

\subsection{Training Details}
% \vspace{-2mm}
% \subsubsection{Training Details}
% \vspace{-2mm}

% we track the number of synthetic queries $Q$ required as a proxy for measuring the likely real human effort involved, which is crucial for methods that rely on human-in-the-loop preference feedback. 
We trained policies and rendered videos on a single A100 GPU machine. The total time for a full experiment was less than one day of wall clock time. We utilized GPT-4, specifically GPT-4-0613, as the backbone LLM in the Proxy Human Preference experiment. For the Human-in-the-loop Preference experiment, we employ GPT-4o.

% \vspace{-2mm}
% \subsubsection{}
% \vspace{-2mm}

% Eureka can access the ground-truth task score and select the highest RTS from  as the  for each experiment. 
% For a fair comparison,  Unlike \citep{ma2023eureka}, which additionally conducts 5 independent PPO training runs for the reward function with the highest RTS and reports the average of 5 maximum task metric values sampled at fixed intervals as the task score, we focus on selecting the best reward function according to human preferences, yielding policies better aligned with the task description. 

\vspace{-2mm}
\subsection{Results of Proxy Human Preference}
% \textbf{Experiment Details.}
\textbf{Experiment Setup.}
In {\name}, we use human-designed sparse rewards as proxies to simulate ideal human preferences. Specifically, in each iteration, we select the reward function with the highest RTS as the good example and the reward function with the lowest RTS as the bad example for feedback. All baseline methods leverage dense rewards to simulate proxy human preference, offering a stronger and more informative signal for labeling preferences. If the cumulative reward of trajectory 1 is greater than that of trajectory 2, then trajectory 1 is preferred over trajectory 2. We also tried sparse rewards as proxy human preference in baseline methods and observed similar performance, shown in Appendix \ref{app: add}.

% For {\name} and baselines, we track the number of synthetic queries $Q$ required as a proxy for measuring the likely real human effort involved, which is crucial for methods that rely on human-in-the-loop preference feedback. Specifically, we define a single query as a human comparing two trajectories and providing a preference. In {\name}, each iteration generates $K$ reward function samples, resulting in $K$ corresponding videos. The human compares these videos, first selecting the best one, then picking the worst from the remaining $K-1$ videos. After $N=5$ iterations, the best video of each iteration is compared to select the overall best. The number of human queries $Q$ can be calculated as $Q = (K-1) \times 2N - 1 $. For {\name}, with $K=6$ and $N=5$, this results in $Q=49$. 

% In baselines, the simulated human teacher compares two sampled trajectories and provides a preference label to update the reward model. We set the maximum number of queries to $Q=49$, matching {\name}, and also test $Q=15k$, denoted as Baseline-\#$Q$ in Table \ref{tab: proxy}, to compare the final task score (FTS) across different tasks. 

\textbf{Main Results.}
\begin{table*}
% \vspace{-4mm}
\caption{The final task score of all methods across different tasks in IssacGym.
The top result and those within one standard deviation are highlighted in bold. Standard deviations are provided in Table \ref{tab: app-proxy} of Appendix \ref{app: add} due to space limitations.}
\centering
% \vspace{-2mm}

\resizebox{1.0\textwidth}{!}{
\begin{tabular}{cccccccccc}
\toprule
             & Cart. & Ball. & Quad. & Anymal  & Ant    & Human. & Franka & Shadow & Allegro  \\
\midrule
PrefPPO-49   & \textbf{499}      & \textbf{499}         & -1.066     & -1.861  & 0.743  & 0.457    & 0.0044 & 0.0746 & 0.0125   \\
PEBBLE-49  & \textbf{499}      & \textbf{499}         & -1.191     & -1.3357  & 5.9891  & 3.67    & 0.0453 & 0.2627 & 0.1467   \\
SURF-49  & \textbf{499}      & \textbf{499}         & -1.202    & -1.35  & 0.874  & 2.406    & 0.0345 & 0.2338 & 0.2002   \\
\hline
PrefPPO-15k  & \textbf{499}      & \textbf{499}         & -0.250     & -1.357  & 4.626  & 1.317    & 0.0399 & 0.0468 & 0.0157   \\
PEBBLE-15k  & \textbf{499}      & \textbf{499}         & -0.231     & -0.730  & 8.543  & 6.162    & 0.8613 & 0.246 & 0.2755   \\
SURF-15k  & \textbf{499}      & \textbf{499}         & -0.266     & -0.76  & 7.859  & 3.532    & 0.5466 & 0.3199 & 0.2352  \\
\hline
{\name}(Ours)   & \textbf{499}      & \textbf{499}         & \textbf{-0.0195}    & \textbf{-0.007} & \textbf{12.04} & \textbf{9.227}    & \textbf{0.9999} & \textbf{13.231} & \textbf{25.030}  \\
\hline
\hline
Eureka       & 499     & 499         & -0.023     & -0.003 & 10.86 & 9.059    & 0.9999 & 11.532 & 25.250   \\
\bottomrule
\end{tabular}
}
\label{tab: proxy}
% \vspace{-3mm}
\end{table*} Table \ref{tab: proxy} shows the final task score (FTS) for {\name} and baseline methods with $Q=49, 15k$ across IsaacGym tasks. Additional results with $Q=150, 1.5k$ can be found in Table \ref{tab: app-proxy} of Appendix \ref{app: add}.
As shown in Table \ref{tab: proxy}, for the simpler tasks like \textit{Cartpole} and \textit{BallBalance}, all methods achieve equal performance. Notably, we observe that for these particularly simple tasks, {\name} can generate correct reward functions in a zero-shot manner, without requiring feedback. As a result, {\name} only requires querying the human 5 times, while baseline methods, after 5 queries, fail to train a reasonable reward model with the preference-labeled data.
For relatively more challenging tasks, Baseline-49 performs significantly worse than {\name} when using the same number of human queries. In fact, Baseline-49 fails in most tasks. As the number of human queries increases, baselines' performance improves across most tasks, but it still falls noticeably short compared to {\name}. This demonstrates that {\name}, with the integration of LLMs, can reduce human effort in preference-based learning by at least 30 times.

% \begin{figure}
% 	\centering
% 	% \vspace{-6mm}
% 	% \hspace{-2mm}
% 		\includegraphics[width=0.35\textwidth]{figures/count.png}
% 	\centering 
% 	\vspace{-4mm}
%     \caption{Distribution of which iteration is selected as the top-scoring iteration. While it is not perfectly monotonic, we observe that the final iteration is generally the best one, suggesting that the inferred reward is gradually approaching the ground-truth reward.}
% 	\vspace{-4mm}
%     \label{fig: fts}
% \end{figure}
\textbf{Performance Analysis.}
We further report the performance of reward-generation methods that utilize ground-truth sparse rewards, which serve as an approximate upper bound on the expected performance {\name} could achieve. For this, we use Eureka~\citep{ma2023eureka}, a state-of-the-art LLM-powered reward design method that leverages sparse rewards as fitness scores. Specifically, in Eureka, the reward function with the highest RTS is selected as the candidate reward function for feedback in each iteration. Additionally, RTS is incorporated as the ``task score'' in the reward reflection prompt sent to the LLM. Original Eureka generates 16 reward functions in each iteration without checking their executability, assuming at least one will typically work across all considered environments in the first iteration. To ensure a fair comparison, we modified Eureka to generate a fixed number of executable reward functions, specifically $K=6$ per iteration, the same as {\name}. This adjustment improves Eureka’s performance in more challenging tasks, where it often generates fewer executable reward functions. As shown in Table \ref{tab: proxy}, {\name} surprisingly achieves comparable performance, indicating that {\name}’s use of LLMs for preference learning is effective.

% From the analysis conducted across 7 tasks where zero-shot generation of optimal reward functions was not feasible in the first iteration, we examined which iteration's RTS was chosen as the final FTS. The distribution of RTS selections over iterations is illustrated in Fig.~\ref{fig: fts}. The results indicate that FTS selections do not always come from the last iteration; some are also derived from earlier iterations. However, the majority of FTS selections originate from iterations 4 and 5, suggesting that {\name} is progressively refining and enhancing the reward functions over successive iterations as opposed to randomly generating diverse reward functions.

% \vspace{-2mm}
\subsection{Method Analysis}
% \vspace{-2mm}
\begin{table*}
% \vspace{-4mm}
\caption{Ablation studies on {\name} modules. The runs have fairly high variance so we highlight the top two results in bold. The full table with std. deviations included can be found in Appendix \ref{app: add}. We observe that {\name} with all of the components is consistently the best performing, suggesting that most of the components are useful.}
\centering
% \vspace{-2mm}
\resizebox{1.0\textwidth}{!}{
\begin{tabular}{cccccccccc}
\toprule
              & Cart. & Ball. & Quad. & Anymal  & Ant    & Human. & Franka & Shadow & Allegro  \\
\midrule
{\name} w/o RT   & \textbf{499}      & \textbf{499}         & -0.0340    & -0.387  & 10.50 & 8.337    & \textbf{0.9999} & 10.769 & \textbf{25.641}   \\
{\name} w/o RTD  & \textbf{499}      & \textbf{499}         & -0.0216    & \textbf{-0.009}  & 10.53 & \textbf{9.419}    & \textbf{1.0000} & 11.633 & 23.744   \\
{\name} w/o RTDB & \textbf{499}      & \textbf{499}         & \textbf{-0.0136}    & -0.014  & \textbf{11.97} & 8.214    & 0.5129 & \textbf{13.663} & \textbf{25.386}   \\
OpenLoop  & \textbf{499}      & \textbf{499}         & -0.0410    & -0.016  & 9.350  & 8.306    & \textbf{0.9999} & 9.476  & 23.876   \\
{\name}(Ours)    & \textbf{499}      & \textbf{499}         & \textbf{-0.0195}    & \textbf{-0.007} & \textbf{12.04} & \textbf{9.227}    & \textbf{0.9999} & \textbf{13.231} & 25.030  \\
\bottomrule
\end{tabular}
}

\vspace{-2mm}
\label{tab: ab}
\end{table*}

To validate the effectiveness of {\name}'s module design, we conducted ablation studies. We aim to answer several questions that could undermine the results presented here:
% \vspace{-2mm}
\begin{enumerate}[itemsep=0pt]
    \item Are components such as the reward trace or the reward difference helpful?
    \item Is the LLM actually performing preference learning? Or is it simply zero-shot outputting the correct reward function due to the task being in the training data?
\end{enumerate}
\vspace{-2mm}
\subsubsection{Ablations}

The results of the ablations are shown in Table \ref{tab: ab}. In these studies, ``{\name} w/o RT'' refers to removing the reward trace from the prompts sent to the LLMs. ``{\name} w/o RTD'' indicates the removal of both the reward trace and the differences between historical reward functions from the prompts. ``{\name} w/o RTDB'' removes the reward trace, differences between historical reward functions, and bad reward functions, leaving only the good reward functions and their evaluation in the prompts. The ``OpenLoop'' configuration samples \( K \times N \) reward functions without any feedback, corresponding to the ability of the LLM to zero-shot accomplish the task. 

Due to the large variance of the experiments (see Appendix), we mark the top two results in bold. As shown, {\name} achieves top 2 results in 8 out of 9 tasks and is comparable on the \textit{Allegro} task. The ``OpenLoop'' configuration performs the worst, indicating that our method does not solely rely on GPT-4’s either having randomly produced the right reward function or having memorized the reward function during its training. This improvement is further demonstrated in Sec. \ref{sec: example}, where we show the step-by-step improvements of ICPL through proxy human preference feedback. Additionally, ``{\name} w/o RT'' underperforms on multiple tasks, highlighting the importance of incorporating the reward trace of historical reward functions into the prompts.

% \ev{I'm going to be really annoying and suggest that the numbers if fig: impro get shifted so that the vertical lines to not pass through them. It looks kinda sloppy.}

\subsubsection{Improvement Analysis} \label{sec: example}

\begin{figure}
	\centering
	% \vspace{-12mm}
	% \hspace{-2mm}
 \includegraphics[width=0.35\textwidth]{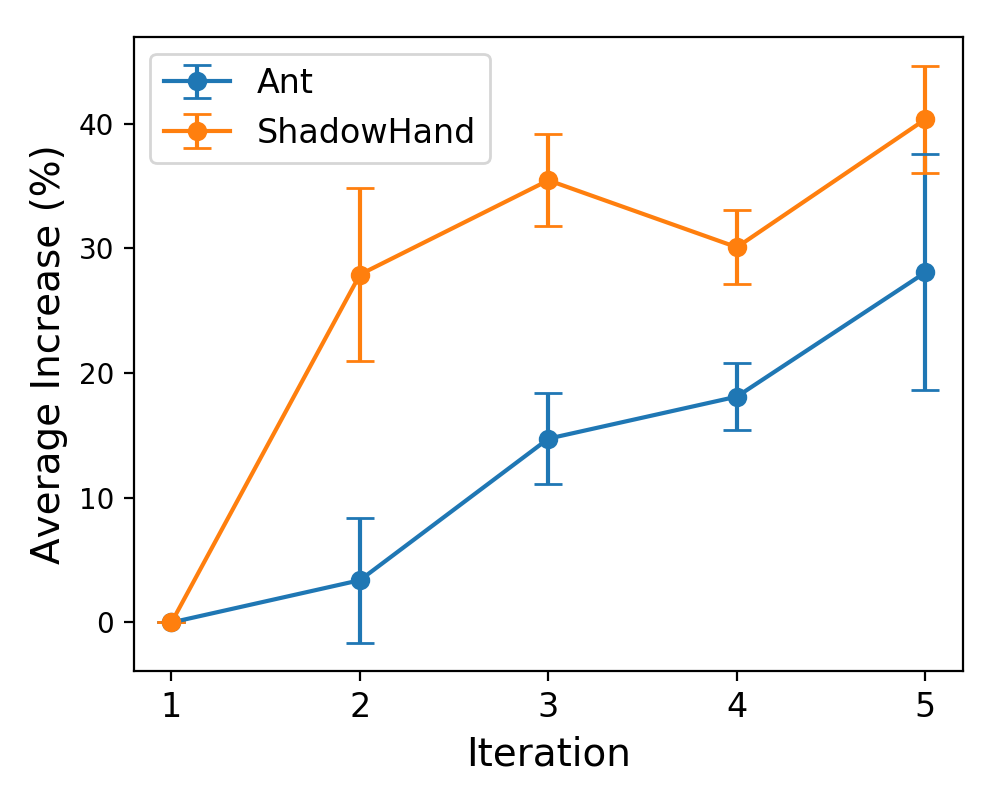}
	\centering 
	\vspace{-4mm}
    \caption{Average improvement of the Reward Task Score (RTS) over successive iterations relative to the first iteration in {\name} for the Ant and ShadowHand tasks, demonstrating the method's effectiveness in refining reward functions.}
	\vspace{-4mm}
    \label{fig: impro}
\end{figure}

Table \ref{tab: proxy} presents the performance achieved by ICPL. While it is possible that the LLMs could generate an optimal reward function in a zero-shot manner, the primary focus of our analysis is not solely on absolute performance values. Rather, we emphasize whether ICPL is capable of enhancing performance through the iterative incorporation of preferences. We calculated the average RTS improvement over iterations relative to the first iteration for the two tasks with the largest improvements compared with ``OpenLoop'', \textit{Ant} and \textit{ShadowHand}. \rebuttal{As shown in Fig. \ref{fig: impro}, the RTS exhibits an upward trend, demonstrating its effectiveness in improving reward functions over time. The individual curves can be found in Appendix \ref{app: add}.}
% We note that this trend is roughly monotonic, indicating that on average the LLM is using the preferences to construct reward functions that are closer to the ground-truth reward.
We further use an example in the \textit{Humanoid} task to demonstrate how {\name} progressively generated improved reward functions over successive iterations in Appendix \ref{app: imp}.

\subsection{Results of Human-in-the-loop Preference}

To address the limitations of proxy human preferences, which simulate idealized human preference and may not fully capture the challenges humans may face in providing preferences, we conducted experiments with real human participants. We recruited 7 volunteers for human-in-the-loop experiments, with 5 assigned to IsaacGym tasks and 2 to a newly designed task. Additionally, 20 volunteers were recruited to evaluate the performance of different methods. None of the volunteers had prior experience with these tasks, ensuring an unbiased evaluation based on their preferences.

% \vspace{-2mm}
% \subsubsection{Human experiment setup}
% \vspace{-2mm}
Before the experiment, each volunteer was provided with a detailed explanation of the experiment’s purpose and process. 
% Clear instructions were given to minimize confusion during the decision-making process \ev{Point out that the instructions are in the appendix? Otherwise, `clear` is debatable.}.
Additionally, volunteers were fully informed of their rights, and written consent was obtained from each participant. The experimental procedure was approved by the department's ethics committee to ensure compliance with institutional guidelines on human subject research. \rebuttal{The detailed setup can be found in Appendix \ref{app: setup}.}

% \vspace{-2mm}
\subsubsection{IsaacGym Tasks}
% \vspace{-2mm}
Due to the simplicity of the \textit{Cartpole}, \textit{BallBalance}, \textit{Franka} tasks, where LLMs were able to zero-shot generate correct reward functions without any feedback, these tasks were excluded from the human trials. The \textit{Anymal} task, which involved commanding a robotic dog to follow random commands, was also excluded as it was difficult for humans to evaluate whether the commands were followed based solely on the videos. For the 5 adopted tasks, we describe in the Appendix \ref{sec: app-human} how humans infer tasks through videos and the potential reasons that may lead to preference rankings that do not accurately reflect the task.
% For the \textit{ShadowHand} and \textit{AllegroHand} tasks, where the objective was to spin an object to a target orientation, humans inferred the quality of the policy based on the continuous spinning of the object.
\begin{table*}
% \vspace{-4mm}
\centering
\caption{The final task score of human-in-the-loop preference across 5 IsaacGym tasks. The values in parentheses represent the standard deviation.}
\label{tab: real}
\vspace{-2mm}
\begin{tabular}{ccccccc}
\toprule
           & Quadcopter & Ant    & Humanoid & Shadow & Allegro   \\
\midrule
OpenLoop  & -0.0410 (0.32)  & 9.350 (2.35)  & 8.306 (1.63) & 9.476 (2.44) &  23.876 (7.91)  \\
{\name}-proxy & -0.0195(0.09)    & 12.040 (1.69) & 9.227 (0.93)  & 13.231 (1.88) & 25.030 (3.72)  \\
{\name}-real  & -0.0183 (0.29)   & 11.142 (0.37) & 8.392 (0.53) & 10.74 (0.92)& 24.134 (6.52)\\
\bottomrule
\end{tabular}

% \vspace{-4mm}
\end{table*}

Table \ref{tab: real} presents the FTS for the human-in-the-loop preference experiments conducted across 5 suitable IsaacGym tasks, labeled as ``ICPL-real''. The results of the proxy human preference experiment are labeled as ``ICPL-proxy''. As observed, the performance of ``ICPL-real'' is comparable or slightly lower than that of ``ICPL-proxy'' in all 5 tasks, yet it still outperforms the ``OpenLoop'' results in 3 out of 5 tasks. This indicates that while humans may have difficulty providing consistent preferences from videos as proxies, their feedback can still be effective in improving performance when combined with LLMs.

\subsubsection{HumanoidJump Task}

% In , there is no reward can serve as task metric to provide quantitative results. instead, we use human vote to provide quantitative results of  so we  Defining a precise task metric for this objective is challenging, as the criteria for human-like jumping are not easily quantifiable. 
\begin{figure}[ht]
	\centering
	% \vspace{-6mm}
	% \hspace{-2mm}
		\includegraphics[width=0.6\linewidth]{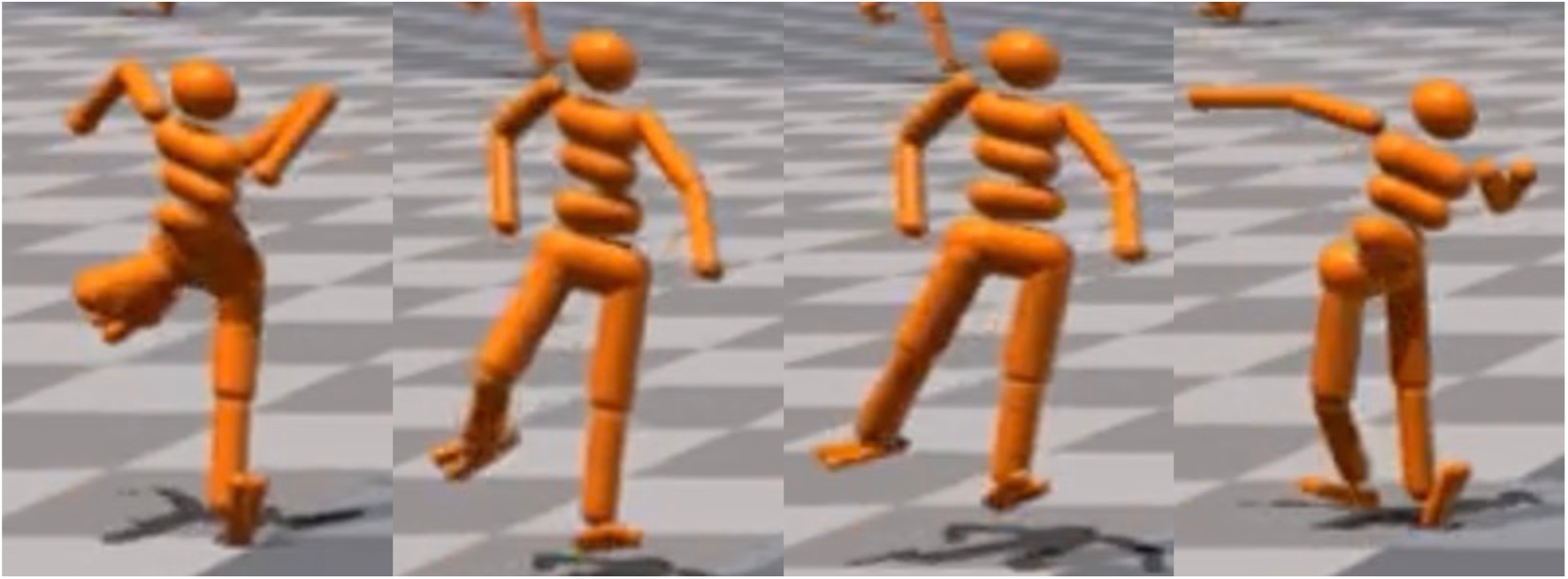}
	\centering 
	% \vspace{-6mm}
    \caption{A common behavior.}
	\vspace{-4mm}
    \label{fig: iter}
\end{figure}
The task-specific prompts used in the newly designed \textit{HumanoidJump} task are detailed in Appendix \ref{app: jump}. The most common behavior observed in this task, as illustrated in Fig.~\ref{fig: iter}, is what we refer to as the ``leg-lift jump.'' This behavior involves initially lifting one leg to raise the center of mass, followed by the opposite leg pushing off the ground to achieve lift. The previously lifted leg is then lowered to extend airtime. Various adjustments of the center of mass with the lifted leg were also noted. This behavior meets the minimal metric of a jump: achieving a certain distance off the ground.
If feedback were provided based solely on this minimal metric, the ``leg-lift jump'' would likely be selected as a candidate reward function. However, 
% our experiments demonstrate that 
such candidates show limited improvement in subsequent iterations, failing to evolve into more human-like jumping behaviors.

\begin{figure}[ht]
    \centering
    % \vspace{-2mm}
    \includegraphics[width=1.0\linewidth]{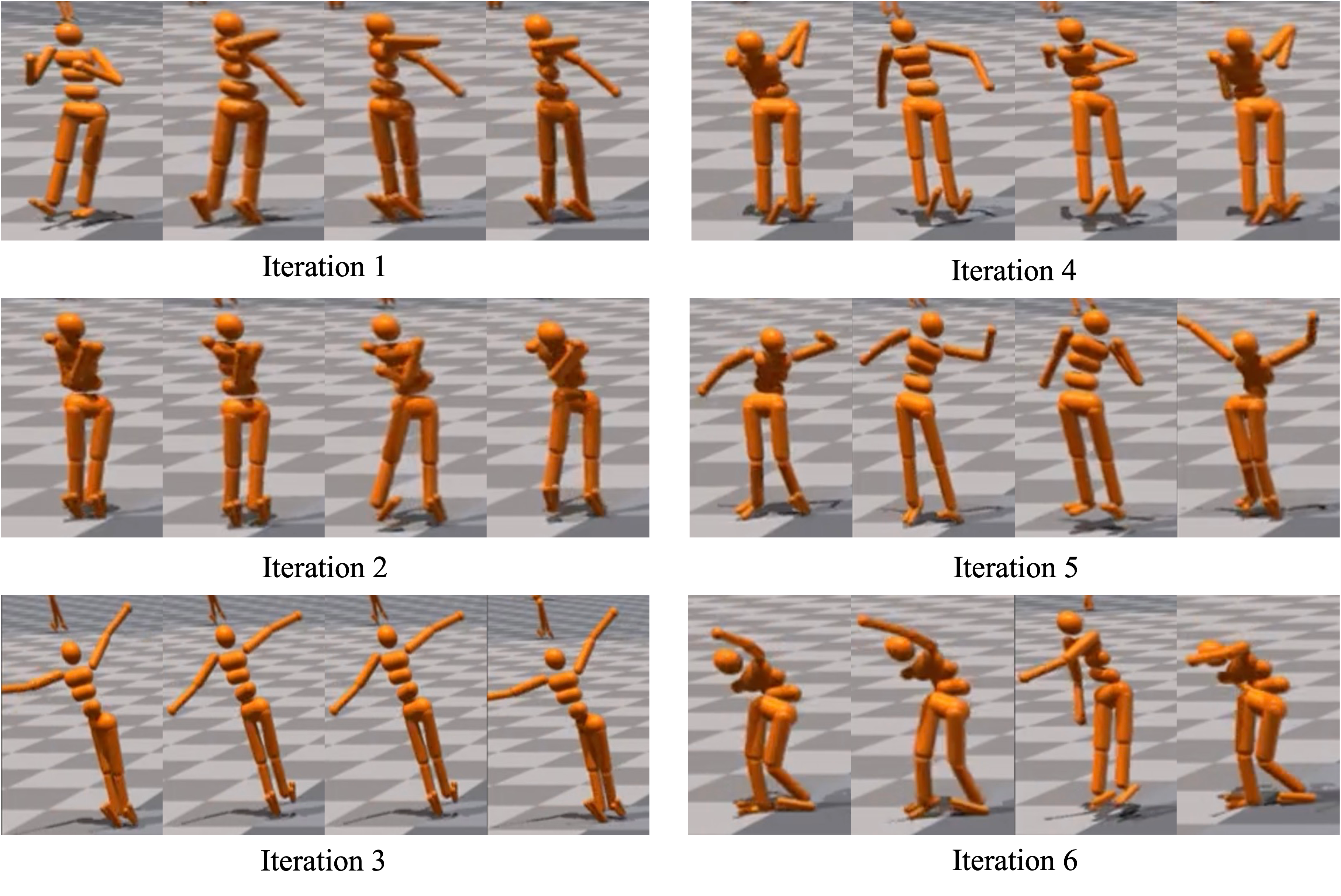}
    \vspace{-5mm}
    \caption{The humanoid learns a human-like jump by bending legs and lowering the upper body to shift the center of mass in a trial of human-in-the-loop experiments. Note that both legs are used to jump and the agent bends at the hips.}
    \label{fig: jump}
\vspace{-4mm}
\end{figure}
Conversely, when real human preferences were used to guide the task, the results were notably different. The volunteer judged the overall quality of the humanoid's jump behavior instead of just the metric of leaving the ground. Fig.~\ref{fig: jump} illustrates that the volunteer successfully guided the humanoid towards a more human-like jump by selecting behaviors that, while initially not optimal, displayed promising movement patterns. The reward functions are shown in Appendix \ref{app: reward}.
In the first iteration, ``leg-lift jump'' was not selected despite the humanoid jumping off the ground. Instead, a video where the humanoid appears to attempt a jump using both legs, without leaving the ground, was chosen. By the fifth and sixth iterations, the humanoid demonstrated more sophisticated behaviors, such as bending both legs and lowering the upper body to shift the center of mass, behaviors that are much more akin to a real human jump. The videos can be found at our website.
\begin{table}[ht]%{r}{0.40\linewidth}
\centering
% \vspace{-2mm}
\begin{tabular}{cc}
\toprule
\textbf{Method} & \textbf{Vote} \\ \midrule
OpenLoop        & 3/20                      \\ 
ICPL            & 17/20                     \\ \bottomrule
\end{tabular}
\caption{Human Preferences over different behaviors.}
\label{tab:human_preference}
\vspace{-5mm}
\end{table}

% By the sixth iteration, the behavior became more pronounced. 

\textbf{Quantitative Evaluation.} Since designing a reward metric for the \textit{HumanoidJump} task is challenging, we adopt human votes for quantitative evaluation instead. As a baseline, we use the ``OpenLoop'' configuration, which generates \( K \times N \) reward functions without any feedback, on the HumanoidJump task. In this configuration, we performed 5 independent experiments, each comprising 6 iterations with 6 samples per iteration. A volunteer selected the most preferred video as the final result. 20 additional volunteers were recruited to compare the performance of ICPL and OpenLoop. Each volunteer indicated their preference between two videos presented in random order—one generated by ICPL and the other by OpenLoop. As shown in Table~\ref{tab:human_preference}, 17 out of 20 participants preferred the ICPL agent, demonstrating that ICPL produces behaviors more aligned with human preferences.

% \vspace{-2mm}
\section{Conclusion}\label{sec:conclu}
% \vspace{-2mm}
% Our proposed method, 
% {\name} demonstrates significant potential for improving the human query efficiency of preference-based RL through the integration of LLMs. 
% We demonstrate that LLMs have native preference-learning capabilities. 
By leveraging the generative capabilities of LLMs to autonomously produce reward functions, and iteratively refining them using human preference feedback, {\name} reduces the complexity and human effort typically associated with PbRL. Our experimental results, both in proxy human preference and human-in-the-loop settings, show that {\name} not only surpasses traditional PbRL baselines in human query efficiency but also competes effectively with methods utilizing ground-truth rewards instead of preferences. Furthermore, the success of {\name} in complex, subjective tasks like humanoid jumping highlights its versatility in capturing nuanced human intentions, opening new possibilities for future applications in complex real-world scenarios where traditional reward functions are difficult to define.

\textbf{Limitations.} While {\name} demonstrates significant potential, it faces limitations in tasks where human evaluators struggle to assess performance from video alone, such as \textit{Anymal}'s "follow random commands." In such cases, subjective human preferences may not provide adequate guidance. Future work will explore integrating partially defined metrics with human preferences. Another area for improvement is incorporating text feedback, where participants explain their preferences, potentially guiding the LLM more efficiently. Additionally, we observe that the performance of the task is qualitatively dependent on the diversity of the initial reward functions that seed the search. Relying on the LLM to provide this initial diversity is a current limitation. Furthermore, the limited number of participants in human-in-the-loop experiments may restrict the generalizability of our findings, as it might not fully capture the broad range of human preferences. Another limitation of {\name} is that each iteration involves training new RL policies, resulting in a waiting period of several hours for participants before they can provide additional feedback. This could be addressed by continuously training an RL agent under non-stationary reward functions, which presents a promising direction for future work.

\vspace{-2mm}
\section*{Impact Statement}
This work tackles tasks where clear reward signals are absent, including both sparse and shaped rewards in RL training. We introduce a novel preference-based RL method that improves sample efficiency through LLM guidance. The study complies with ethical standards: the experimental procedure was approved by the department's ethics committee, and participants were informed of their rights. This paper presents work whose goal is to advance the field of  Machine Learning. There are many potential societal consequences of our work, none which we feel must be specifically highlighted here.

% Authors are \textbf{required} to include a statement of the potential 
% broader impact of their work, including its ethical aspects and future 
% societal consequences. This statement should be in an unnumbered 
% section at the end of the paper (co-located with Acknowledgements -- 
% the two may appear in either order, but both must be before References), 
% and does not count toward the paper page limit. In many cases, where 
% the ethical impacts and expected societal implications are those that 
% are well established when advancing the field of Machine Learning, 
% substantial discussion is not required, and a simple statement such 
% as the following will suffice:

% ``This paper presents work whose goal is to advance the field of 
% Machine Learning. There are many potential societal consequences 
% of our work, none which we feel must be specifically highlighted here.''

% The above statement can be used verbatim in such cases, but we 
% encourage authors to think about whether there is content which does 
% warrant further discussion, as this statement will be apparent if the 
% paper is later flagged for ethics review.

% In the unusual situation where you want a paper to appear in the
% references without citing it in the main text, use \nocite
\nocite{langley00}

\bibliography{reference}
\bibliographystyle{icml2025}

%%%%%%%%%%%%%%%%%%%%%%%%%%%%%%%%%%%%%%%%%%%%%%%%%%%%%%%%%%%%%%%%%%%%%%%%%%%%%%%
%%%%%%%%%%%%%%%%%%%%%%%%%%%%%%%%%%%%%%%%%%%%%%%%%%%%%%%%%%%%%%%%%%%%%%%%%%%%%%%
% APPENDIX
%%%%%%%%%%%%%%%%%%%%%%%%%%%%%%%%%%%%%%%%%%%%%%%%%%%%%%%%%%%%%%%%%%%%%%%%%%%%%%%
%%%%%%%%%%%%%%%%%%%%%%%%%%%%%%%%%%%%%%%%%%%%%%%%%%%%%%%%%%%%%%%%%%%%%%%%%%%%%%%
\newpage
\appendix
\onecolumn

% Set up the code style
\lstset{
  basicstyle=\ttfamily\footnotesize,
  keywordstyle=\bfseries\color{blue},
  commentstyle=\itshape\color{green!60!black},
  stringstyle=\color{orange},
  showstringspaces=false,
  breaklines=true,
  % frame=single,
  numbers=left,
  numberstyle=\tiny\color{gray},
  tabsize=4
}

We would suggest visiting \url{https://sites.google.com/view/few-shot-icpl/home} for more information and videos.

\section{Full Prompts}
\label{app: prompt}
The prompts used in {\name} for synthesizing reward functions are presented in Prompts \ref{prompt: 1}, \ref{prompt: 2}, and \ref{prompt: 3}. The prompt for generating the differences between various reward functions is shown in Prompt \ref{prompt: 4}.
\definecolor{codegray}{rgb}{0.5,0.5,0.5}
\definecolor{backcolour}{RGB}{245,245,245}

\lstdefinestyle{mystyle}{
    backgroundcolor=\color{backcolour},   
    commentstyle=\color{magenta},
    keywordstyle=\color{blue},
    numberstyle=\tiny\color{codegray},
    numbers=none,
    basicstyle=\fontfamily{\ttdefault}
    \footnotesize,
    breakatwhitespace=false,         
    breaklines=true,                 
    keepspaces=true,    
    frame=single,
    numbersep=5pt,                  
    showspaces=false,                
    showstringspaces=false,
    showtabs=false,                  
    tabsize=2,
    classoffset=1, %
    keywordstyle=\color{violet},
    classoffset=0,
}
\lstset{style=mystyle}
\renewcommand\lstlistingname{Prompt}

\begin{lstlisting}[basicstyle=\fontfamily{\ttdefault}\scriptsize, breaklines=true,caption={Initial System Prompts of Synthesizing Reward Functions}\label{prompt: 1}]
You are a reward engineer trying to write reward functions to solve reinforcement learning tasks as effective as possible.
Your goal is to write a reward function for the environment that will help the agent learn the task described in text. 
Your reward function should use useful variables from the environment as inputs. As an example, the reward function signature can be: 
@torch.jit.script
def compute_reward(object_pos: torch.Tensor, goal_pos: torch.Tensor) -> Tuple[torch.Tensor, Dict[str, torch.Tensor]]:
    ...
    return reward, {}
Since the reward function will be decorated with @torch.jit.script, please make sure that the code is compatible with TorchScript (e.g., use torch tensor instead of numpy array). 
Make sure any new tensor or variable you introduce is on the same device as the input tensors. 
\end{lstlisting}
\begin{lstlisting}[basicstyle=\fontfamily{\ttdefault}\scriptsize, breaklines=true,caption={Feedback Prompts}\label{prompt: 2}]
The reward function has been iterated {current_iteration} rounds. 
In each iteration, a good reward function and a bad reward function are generated. 
The good reward function generated in the x-th iteration is denoted as "iterx-good", and the bad reward function generated is denoted as "iterx-bad". 
The following outlines the differences between these reward functions.

We trained an RL policy using iter1-good reward function code and tracked the values of the individual components in the reward function after every {epoch_freq} epochs and the maximum, mean, minimum values encountered:
<REWARD FEEDBACK>

The difference between iter2-good and iter1-good is: <DIFFERENCE>

<REPEAT UNTIL THE CURRENT ITERATION>

Next, the two reward functions generated in the {current_iteration_ordinal} iteration are provided.
The 1st generated reward function is as follows:
<REWARD FUNCTION>
We trained an RL policy using the 1st reward function code and tracked the values of the individual components in the reward function after every {epoch_freq} epochs and the maximum, mean, minimum values encountered:
<REWARD FEEDBACK>

The 2nd generated reward function is as follows:
<REWARD FUNCTION>
We trained an RL policy using the 2nd reward function code and tracked the values of the individual components in the reward function after every {epoch_freq} epochs and the maximum, mean, minimum values encountered:
<REWARD FEEDBACK>

The following content is the most important information.
Good example: 1st reward function. Bad example: 2nd reward function.
You need to modify based on the good example. DO NOT based on the code of the bad example.
Please carefully analyze the policy feedback and provide a new, improved reward function that can better solve the task. Some helpful tips for analyzing the policy feedback:
    (1) If the values for a certain reward component are near identical throughout, then this means RL is not able to optimize this component as it is written. You may consider
        (a) Changing its scale or the value of its temperature parameter
        (b) Re-writing the reward component 
        (c) Discarding the reward component
    (2) If some reward components' magnitude is significantly larger, then you must re-scale its value to a proper range
Please analyze each existing reward component in the suggested manner above first, and then write the reward function code. 
\end{lstlisting}
\begin{lstlisting}[basicstyle=\fontfamily{\ttdefault}\scriptsize, breaklines=true,caption={Prompts of Tips for Writing Reward Functions}\label{prompt: 3}]
The output of the reward function should consist of two items:
    (1) the total reward,
    (2) a dictionary of each individual reward component.
The code output should be formatted as a python code string: "```python ... ```".

Some helpful tips for writing the reward function code:
    (1) You may find it helpful to normalize the reward to a fixed range by applying transformations like torch.exp to the overall reward or its components
    (2) If you choose to transform a reward component, then you must also introduce a temperature parameter inside the transformation function; this parameter must be a named variable in the reward function and it must not be an input variable. Each transformed reward component should have its own temperature variable
    (3) Make sure the type of each input variable is correctly specified; a float input variable should not be specified as torch.Tensor
    (4) Most importantly, the reward code's input variables must contain only attributes of the provided environment class definition (namely, variables that have prefix self.). Under no circumstance can you introduce new input variables.
\end{lstlisting}
\begin{lstlisting}[basicstyle=\fontfamily{\ttdefault}\scriptsize, breaklines=true,caption={Prompts of Describing Differences}\label{prompt: 4}]
You are an engineer skilled at comparing the differences between two reward function code snippets used in reinforcement learning.
Your goal is to describe the differences between two reward function code snippets.
The following are two reward functions written in Python code used for the task: 
<TASK_DESCRIPTION>
The first reward function is as follows: 
<REWARD_FUNCTION>
The second reward function is as follows: 
<REWARD_FUNCTION>
Please directly describe the differences between these two codes. No additional descriptions other than the differences are required.
\end{lstlisting}

\section{{\name} Details}
The full pseudocode of {\name} is listed in Algo. \ref{algo: app-icpl}. 
\rebuttal{We provide an example to further explain the reward components of the reward function. Take the Humanoid task as an example, where the goal is to make the humanoid run as fast as possible. Below is a typical set of reward components generated by ICPL.}
\rebuttal{\begin{itemize}
    \item \texttt{velocity\_reward}: reward for forward velocity (run fast)
    \item \texttt{upright\_reward}: encouragement for maintaining upright posture
    \item \texttt{force\_penalty}: penalize high force usage (energy efficiency)
    \item \texttt{unnatural\_pose\_penalty}: penalize unnatural joint angles
    \item \texttt{action\_penalty}: penalize large actions (for smoother movement)
\end{itemize}}
\rebuttal{The total reward is the sum of these individual components. Designing such a reward requires specifying and balancing five different aspects of behavior, which is likely nontrivial.
}

\section{Baseline Details}
\label{app: baseline}

\subsection{PrefPPO}
The baseline PrefPPO adopted in our experiments comprises two primary components: agent learning and reward learning, as outlined in \citet{lee2021bpref}.  Algo. \ref{algo: PrefPPO} illustrates the pseudocode for PrefPPO.  Throughout this process, the method maintains a policy denoted as $\pi_{\varphi}$ and a reward model represented by $\hat{r_{\psi}}$.

\textbf{Agent Learning}. In the agent learning phase, the agent interacts with the environment and collects experiences. The policy is subsequently trained using reinforcement learning, to maximize the cumulative rewards provided by the reward model $\hat{r_{\psi}}$. We utilize the on-policy reinforcement learning algorithm PPO~\citep{schulman2017proximalpolicyoptimizationalgorithms} as the backbone algorithm for training the policy. Additionally, we apply unsupervised pre-training to match the performance of the original benchmark. Specifically, during earlier iterations, when the reward model has not collected sufficient trajectories and exhibits limited progress, we utilize the state entropy of the observations, defined as $H(s)=-\mathbb{E}_{s\sim p(s)}[\log p(s)]$, as the goal for agent training.
During this process, trajectories of varying lengths are collected. Formally, a trajectory $\sigma$ is defined as a sequence of observations and actions ${(s_1,a_1),\dots,(s_t,a_t)}$ that represents the complete interaction of the agent with the environment, concluding at timestep 
$t$.

\textbf{Reward Learning}. A preference predictor is developed using the current reward model to align with human preferences, formulated as follows:
\begin{equation}
    P_{\psi}[\sigma^1\succ \sigma^0]=\frac{\exp \left(\sum_t \hat{r_{\psi}}(s_t^1,a_t^1)\right)}{\sum_{i \in \{0,1\}} \exp \left(\sum_t \hat{r_{\psi}}(s_t^i,a_t^i)\right)},
\end{equation}
where $\sigma_0 = {(s_1^0,a_1^0),\dots,(s_{l_0}^0,a_{l_0}^0)}$ and $\sigma_1 = {(s_1^1,a_1^1),\dots,(s_{l_1}^1,a_{l_1}^1)}$ represent two complete trajectories with different trajectory length $l_0$ and $l_1$. $P_{\psi}[\sigma^1 \succ \sigma^0]$ denotes the probability that trajectory $\sigma^1$ is preferred over $\sigma^0$ as indicated by the preference predictor. In the original PrefPPO framework, test task trajectories are of fixed length, allowing for the extraction of fixed-length segments to train the reward model. However, the tasks in this paper have varying trajectory lengths, so we use full trajectory pairs as training data instead of segments. We also tried zero-padding trajectories to the maximum episode length and then segmenting them, but this approach was ineffective in practice.

To provide more effective labels, the original PrefPPO utilizes dense rewards $r$ to simulate oracle human preferences, which is 
\begin{equation}
    P[\sigma^1\succ \sigma^0]=\begin{cases}
    1& \text{If } \sum_t r(s_t^1,a_t^1)>\sum_t r(s_t^1,a_t^1)\\
    0& \text{Otherwise}
    \end{cases}.
\end{equation}

 The probability $P[\sigma^1 \succ \sigma^0]$ reflects the preference of the ideal teacher, which is perfectly rational and deterministic, without incorporating noise. We utilize the default dense rewards in the adopted IsaacGym tasks, which differ from {\name} that use sparse rewards (task metrics) as the proxy preference. While we also experimented with sparse rewards in PrefPPO and found similar performance (refer to Table \ref{tab: app-sparse}), we opted to retain the original PrefPPO approach in all experiments. The reward model is trained by minimizing the cross-entropy loss between the predictor and labels, utilizing trajectories sampled from the agent learning process. Note that since the agent learning process requires significantly more experiences for training than reward training, we only use trajectories from a subset of the environments for reward training.

To sample trajectories for reward learning, we employ the disagreement sampling scheme from \citet{lee2021bpref} to enhance the training process. This scheme first generates a larger batch of trajectory pairs uniformly at random and then selects a smaller batch with high variance across an ensemble of preference predictors. The selected pairs are used to update the reward model.

For a fair comparison, we recorded the number of times PrefPPO queried the oracle human simulator to compare two trajectories and obtain labels during the reward learning process, using this as a measure of the human effort involved. In the proxy human experiment, we set the maximum number of human queries $Q$ to $49, 150, 1.5k, 15k$. Once this limit is reached, the reward model ceases to update, and only the policy model is updated via PPO. Algo. \ref{algo: rewardlearning} illustrates the pseudocode for reward learning.

% \rebuttal{
% % In experiments, following the implementation of B-Pref(\citep{lee2021bpref}), we use a three-layer neural network with 256 hidden units each and leaky ReLUs as the reward model. 
% To ensure stability, we use an ensemble of three reward models and bound the output of $[-1,1]$ using tanh function. Each model is trained by minimizing the cross-entropy loss using Adam optimizer~\citep{kingma2017adammethodstochasticoptimization} with a learning rate of 0.0003. Each time we update reward models, we query human's feedback for $10, 10, 32, 128$ samples (\texttt{MbSize} in the pseudocode) when the query limit $Q$ is $49, 150, 1.5k, 15k$, respectively. We set the time interval between two reward training (\texttt{RewardTrainingInterval} in the pseudocode) be $50, 50, 50, 15$ iterations when the query limit $Q$ is $49, 150, 1.5k, 15k$ to improve the stability. This time interval is the same for all the tasks if we convert the training process in each task to a process with $3000$ iterations.}

% \rebuttal{For PPO agents, we follow the parameters in Isaac Gym~\citep{makoviychuk2021isaacgymhighperformance}, such as network units and number of enviorments to ensure its high performances.} 

\subsection{PEBBLE}

 PEBBLE~\citep{lee2021pebblefeedbackefficientinteractivereinforcement} is a popular feedback-efficient preference-based RL algorithm. It improves the feedback efficiency of the algorithm by mainly utilizing two modules: unsupervised pre-training and off-policy learning. The unsupervised pre-training module is introduced in the PrefPPO section, and we also include it in PEBBLE with the same setting. PEBBLE utilizes the off-policy algorithm SAC~\citep{haarnoja2018softactorcriticoffpolicymaximum} instead of PPO as the backbone RL algorithm. SAC stores the agent's past experiences in a replay buffer and reuses these experiences during the training process. PEBBLE relabels all past experiences in the replay buffer every time it updates the reward model.

% \rebuttal{In experiments, we utilize the same reward model with the same hyperparameters as in PrefPPO. We also follow the hyperparameters in PEBBLE~\citep{lee2021pebblefeedbackefficientinteractivereinforcement} for the RL training and adjust some hyperparameters for the isaac environments.  
% These hyperparameters are shown in Table ~\ref{tab:pebble_hyperparameters}.}

\subsection{SURF}
SURF~\citep{park2022surf} is a framework that uses unlabeled samples with data augmentation to improve the efficiency of reward training. In our experiments, the length of trajectories is varied and may affect the evaluation of the trajectories. Therefore, we do not apply the data augmentation technique and only utilize the semi-supervised learning method in SURF.

In addition to the labeled pairs of trajectories $\mathcal{D}_l=\{(\sigma^0_l,\sigma^1_l,y)^i\}_{i=1}^{N_l}$, SURF samples another unlabeled dataset $\mathcal{D}_U=\{(\sigma^0_u,\sigma^1_u)^i\}_{u=1}^{N_u}$ to optimize the reward model. Specifically, during each update of the reward model, SURF not only samples a set of trajectories and queries a human teacher for labels, but also samples additional trajectory pairs. These additional pairs are assigned pseudo-labels generated by the current reward model.
\begin{equation}
    \hat{y_u}(\sigma_u^0,\sigma_u^1)=\begin{cases}
1& \texttt{If }P_{\psi}[\sigma_u^1\succ \sigma_u^0]>0.5.\\
0& \texttt{Otherwise}.
\end{cases}
\end{equation}

Here $\psi$ is the preference predictor based on the current reward model. During the training process of reward model, SURF will also use the unlabeled samples for training if the confidence of the predictor is higher than a pre-defined threshold. In experiments, we follows the implementation of SURF~\citep{park2022surf}.
% Specifically, the training objective for the reward model is 
% $$
% \mathcal{L}_{\text{SSL}} = 
% \mathbb{E}_{\substack{(\sigma_l^0, \sigma_l^1, y) \sim \mathcal{D}_l \\ (\sigma_u^0, \sigma_u^1) \sim \mathcal{D}_u}} \left[ 
% \mathcal{L}_{\text{Reward}}(\sigma_l^0, \sigma_l^1, y) 
% + \lambda \cdot \mathcal{L}_{\text{Reward}}(\sigma_u^0, \sigma_u^1, \hat{y}) \cdot 
% \mathbf{1} \left( P_\psi[\sigma_u^{h^*} \succ \sigma_u^{1 - h^*}] > \tau \right)
% \right]
% $$
% where $\mathcal {L}_{\text{Reward}}(\sigma^0,\sigma^1,y)$ is the cross entropy loss
% $$
% \mathcal {L}_{\text{Reward}}(\sigma^0,\sigma^1,y)=(1-y)\log P_{\psi}[\sigma^0\succ \sigma^1]+y\log P_{\psi}[\sigma^1\succ \sigma^0],
% $$
% $\lambda$ is a hyperparameter for balancing the loss, $\tau$ is a confidence threshold.}

% \rebuttal{ and uses the hyperparameters as shown in Table~\ref{tab:surf_hyperparameters}. We also use the same reward model as PEBBLE. And we use a unlabeled batch 
% ratio $\mu$ which means in each update we sample $\texttt{MbSize}$ labeled samples and $\mu \times \texttt{MbSize}$ unlabeled samples.}

\begin{algorithm}[htbp]

\SetAlgoLined
\SetKwFunction{History}{History} 
\SetKwFunction{Feedback}{Feedback} 
\SetKwProg{Fn}{Function}{:}{}
\KwIn{\# iterations $N$, \# samples in each iterations $K$, environment $\texttt{Env}$, coding LLM $\texttt{LLM}_{RF}$, difference LLM $\texttt{LLM}_{Diff}$}
\Fn{\Feedback{$\texttt{Env}, \texttt{RF}$}}{
    \KwRet The values of each component that make up \texttt{RF} during the training process in \texttt{Env}
}
\Fn{\History{$\texttt{RFlist}, \texttt{Env}, \texttt{LLM}_{Diff}$}}{
    $\texttt{HistoryFeedback} \gets ``"$
    % \texttt{HistoryFeedback} \gets \texttt{Feedback}{(\texttt{Env}, \texttt{RFlist[0]})}$
    
    \For{$i \gets 1$ to \textbf{len}$(\texttt{RFlist})-1$}{
    
    \tcp{The reward trace of historical reward functions}
    $\texttt{HistoryFeedback} \gets \texttt{HistoryFeedback} + \texttt{Feedback}{(\texttt{Env}, \texttt{RFlist[$i-1$]})}$

    \tcp{The differences between historical reward functions}
    $\texttt{HistoryFeedback} \gets \texttt{HistoryFeedback} + \texttt{LLM}_{Diff}(\texttt{DifferencePrompt}+ \texttt{RFlist[$i$]}+ \texttt{RFlist[$i-1$]})$
    }
    \KwRet{$\texttt{HistoryFeedback}$}
}

%\tcp{Use environment as context to initialize prompt}
\tcp{Initialize the prompt containing the environment context and task description}
$\texttt{Prompt} \gets \texttt{InitializePrompt}$

$\texttt{RFlist} \gets []$

\For{$i \gets 1$ to $N$}{
    $\texttt{RF}_1, \dots, \texttt{RF}_K\gets \texttt{LLM}_{RF}(\texttt{Prompt}, K)$
    
    \While{any of $\texttt{RF}_1, \dots, \texttt{RF}_K$ is not executable}
    {$j_1, \dots, j_{K'} \gets$ \text{Index of non-executable reward functions}
    
    \tcp{Regenerate non-executable reward functions}
    $\texttt{RF}_{j_1}, \dots, \texttt{RF}_{j_K'}\gets \texttt{LLM}_{RF}(\texttt{Prompt}, K')$ 
    }
    
    \tcp{Render videos for sampled reward functions}
    $\texttt{Video}_1, \dots, \texttt{Video}_K\gets \texttt{Render}(\texttt{Env}, \texttt{RF}_1), \dots, \texttt{Render}(\texttt{Env}, \texttt{RF}_K)$
    
    \tcp{Human selects the most preferred and least preferred videos}
    $G, B \gets \texttt{Human}(\texttt{Video}_1, \dots, \texttt{Video}_K)$
    
    $\texttt{GoodRF}, \texttt{BadRF} \gets \texttt{RF}_{G}, \texttt{RF}_{B}$
    
    $\texttt{RFlist}.\textbf{append}(\texttt{GoodRF})$

    \tcp{Update prompt for feedback}
    $\texttt{Prompt} \gets \texttt{GoodRF}+\texttt{Feedback}{(\texttt{Env}, \texttt{GoodRF})} + \texttt{BadRF}+\texttt{Feedback}{(\texttt{Env}, \texttt{BadRF})} + \texttt{PreferencePrompt}$

    $\texttt{Prompt} \gets \texttt{Prompt} + \texttt{History}(\texttt{RFlist}, \texttt{Env}, \texttt{LLM}_{Diff})$
}
\caption{\name}
\label{algo: app-icpl}
\end{algorithm}

\begin{algorithm}[H]

\SetAlgoLined
\SetKwFunction{History}{History} 
\SetKwProg{Fn}{Function}{:}{}
\KwIn{\# iterations $B$, \# unsupervised learning iterations $M$, \# rollout steps $S$, reward model $\hat{r_{\psi}}$, \# environments for reward learning $E$,  \# iterations for collecting trajectories \texttt{RewardTrainingInterval},\text{maximal number of human queries} $Q$,  environments $\texttt{Env}$}

\texttt{HumanQueryCount}$\gets 0$

$\texttt{Trajectories} \gets []$

\Fn{\texttt{TrainReward}{($\hat{r_{\psi}}$, \texttt{Trajectories})}}

\Fn{\texttt{CollectRollout}{$(\texttt{RewardType}, S, \texttt{Policy}, \hat{r_{\psi}}, \texttt{Env})$}}{
    $\texttt{RolloutBuffer} \gets []$

    \For{$j \gets 1$ to $S$}{
        $\texttt{Action} \gets \texttt{Policy}(\texttt{Observation})$

        \tcp{Here EnvDones is a binary sequence replied from the envrionment, representing whether the environments are done.}
        $\texttt{NewObservation, EnvReward, EnvDones}\gets \texttt{Env}(\texttt{Actions})$

        \If{\texttt{RewardType} == \texttt{Unsuper}}{
            $\texttt{PredReward}\gets \texttt{ComputeStateEntropy}(\texttt{Observation})$
        }
        \Else{
            $\texttt{PredReward}\gets \hat{r_{\psi}}(\texttt{Observation},\texttt{Action})$
        }
        
        \tcp{Collect trajectories for reward learning}
        $\texttt{Trajectories}\gets \texttt{Trajectories}+(\texttt{Observation},\texttt{Action},\texttt{EnvDones},\texttt{EnvReward})$

        \tcp{Add complete trajectory to reward model}
        \For{$k\gets 1$ to $E$}{
            \If{$\texttt{EnvDones}[\texttt{Env}[k]]$}{
                $\texttt{AddTrajectory}(\hat{r_{\psi}}, \texttt{Trajectories}[k])$

                $\texttt{Trajectories}[k]\gets []$
            }
        }

        \tcp{Reward Learning}
        \If{$j \texttt{ is divisible by RewardTrainingInterval}$ and $\texttt{HumanQueryCount}<Q$}{
            $\hat{r_{\psi}} \gets $$\texttt{TrainReward}(\hat{r_{\psi}}, \texttt{Trajectories})$
        }
        
        \tcp{Collect rollouts for agent learning}
        $\texttt{RolloutBuffer}\gets \texttt{RolloutBuffer}+(\texttt{Observation},\texttt{Action},\texttt{PredReward})$

        $\texttt{Observation}\gets \texttt{NewObservation}$
    }
    \KwRet{\texttt{RolloutBuffer}}
}

$\texttt{Policy}\gets \texttt{Initialize}$

\For{$i \gets 1$ to $B$}{
    \tcp{Collect rollouts and trajectories}
    \If{ $i<M$}{ 
        \texttt{RolloutBuffer} $\gets$ \texttt{CollectRollout}$(\texttt{Unsuper}, S, \texttt{Policy}, \hat{r_{\psi}}, \texttt{Env})$
    }
    \Else{
        \texttt{RolloutBuffer} $\gets$ \texttt{CollectRollout}$(\texttt{RewardModel}, S, \texttt{Policy}, \hat{r_{\psi}}, \texttt{Env})$
    }
    \tcp{Agent Learning: Train agent with the collect RolloutBuffer via PPO, omitted here}
    \texttt{AgentLearning}(\texttt{Policy}, \texttt{RolloutBuffer})
    
}
\caption{PrefPPO}
\label{algo: PrefPPO}
\end{algorithm}

\begin{algorithm}[H]

\SetAlgoLined
\SetKwFunction{History}{History} 
\SetKwProg{Fn}{Function}{:}{}
\KwIn{reward model $\hat{r_{\psi}}$, \# samples for human queries per time \texttt{MbSize}, \# maximal iterations for reward learning \texttt{MaxUpdate}, \text{maximal number of human queries} $Q$, environments $\texttt{Env}$}

\texttt{LabeledQueries}$\gets []$

\texttt{HumanQueryCount}$\gets 0$

\Fn{\texttt{TrainReward}{($\hat{r_{\psi}}$, \texttt{Trajectories})}}{
    \tcp{Use disagreement sampling to sample trajectories}
    $\sigma_0,\sigma_1 \gets \texttt{DisagreementSampling}(\texttt{Trajectories}, \texttt{MbSize})$

    \For{$(x_0,x_1)$ in $(\sigma_0,\sigma_1)$}{
         \tcp{Give oracle human preferences between two trajectories according to the sum of dense reward.}
        \texttt{LabeledQueries}$\gets \texttt{LabeledQueries}+(x_0,x_1,\texttt{HumanQuery}(x_0,x_1))$
         
    \tcp{In experiments, we do not add HumanQueryCount if the pair has already been queried before}

         $\texttt{HumanQueryCount}\gets \texttt{HumanQueryCount}+1$
         
         \If{$\texttt{HumanQueryCount}>Q$}{
            \texttt{BREAK}
        }
    }

    \For{$i \gets 1$ to \texttt{MaxUpdate}}{
        \tcp{Update reward model by minimizing the cross entropy loss and record the accuracy on all pairs.}
        $\hat{r_{\psi}}$, \texttt{Accuracy} $\gets$ \texttt{RewardLearning}($\hat{r_{\psi}}$, \texttt{LabeledQueries})

        \If{\texttt{Accuracy} $\ge 97\%$}{
            \texttt{BREAK}
        }
    }
    \KwRet{$\hat{r_{\psi}}$}
}
\caption{Reward Learning of PrefPPO}
\label{algo: rewardlearning}
\end{algorithm}

\noindent

% feedback is provided directly on full-length trajectories of varying lengths.

% \begin{table}[ht]
% \centering
% \caption{Hyperparameters of PEBBLE.}
% \label{tab:pebble_hyperparameters}
% \begin{tabular}{@{}ll@{\hskip 20pt}ll@{}}
% \toprule
% \textbf{Hyperparameter} & \textbf{Value}         & \textbf{Hyperparameter}       & \textbf{Value}          \\ \midrule
% Initial temperature     & 0.1                   & Hidden units per each layer   & 256 \\
% Optimizer       & Adam (\cite{kingma2017adammethodstochasticoptimization})                    & \# of layers                  & 3     \\
% Learning rate           & 0.001 (Allegro/Shadow Hand),  & Batch Size                    & 4096       \\
%            & 0.0005 (Other tasks),  &                    &        \\
% Critic target update freq & 2                   & Critic EMA $\tau$             & 0.005                            \\
% $(\beta_1, \beta_2)$    & (0.9, 0.999)          & Discount $\gamma$             & 0.99                             \\
%  \bottomrule
% \end{tabular}
% \end{table}

% \rebuttal{
% \begin{table}[h!]
% \centering
% \caption{Hyperparameters of SURF}
% \label{tab:surf_hyperparameters}
% \begin{tabular}{@{}ll@{\hskip 20pt}ll@{}}
% \toprule
% \textbf{Hyperparameter} & \textbf{Value}          \\ \midrule
% Unlabeled batch ratio $\mu$     & 3(when num of queries=15000)                   \\
% & 10(otherwise)\\
% Threshold $\tau$       & 0.99\\
% Loss weight $\lambda$          & 1\\
%  \bottomrule
% \end{tabular}
% \end{table}
% }

\section{Environment Details} \label{app: env}
In Table \ref{tab: app-env}, we present the observation and action dimensions, along with the task description and task metrics for 9 tasks in IsaacGym.

\begin{table}[htbp]
\centering
\begin{tabular}{>{\raggedright\arraybackslash}p{12cm}} 
\toprule
% \multicolumn{1}{c}{\textbf{Tasks for Proxy Human Preference}} \\
% \midrule
\textbf{Environment (obs dim, action dim)} \\
{Task Description} \\
\textit{Task Metric} \\
\hline
\hline
\textbf{Cartpole (4, 1)} \\
To balance a pole on a cart so that the pole stays upright \\
\textit{duration} \\
\hline
\hline
\textbf{Quadcopter (21, 12)} \\
To make the quadcopter reach and hover near a fixed position \\
\textit{-cur\_dist} \\
\hline
\hline
\textbf{FrankaCabinet (23, 9)} \\
To open the cabinet door \\
\textit{1 if cabinet\_pos $>$ 0.39} \\
\hline
\hline
\textbf{Anymal (48, 12)} \\
To make the quadruped follow randomly chosen x, y, and yaw target velocities \\
\textit{-(linvel\_error + angvel\_error)} \\
\hline
\hline
\textbf{BallBalance (48, 12)} \\
To keep the ball on the table top without falling \\
\textit{duration} \\
\hline
\hline
\textbf{Ant (60, 8)} \\
To make the ant run forward as fast as possible \\
\textit{cur\_dist - prev\_dist} \\
\hline
\hline
\textbf{AllegroHand (88, 16)} \\
To make the hand spin the object to a target orientation \\
\textit{number of consecutive successes where 
current success is 1 if rot\_dist $<$ 0.1} \\
\hline
\hline
\textbf{Humanoid (108, 21)} \\
To make the humanoid run as fast as possible \\
\textit{cur\_dist - prev\_dist} \\
\hline
\hline
\textbf{ShadowHand (211, 20)} \\
To make the shadow hand spin the object to a target orientation \\
\textit{number of consecutive successes where 
current success is 1 if rot\_dist $<$ 0.1} \\
\bottomrule
\end{tabular}
\caption{Details of IssacGym Tasks.}
\label{tab: app-env}
\end{table}

\begin{table}
\centering
\resizebox{1.0\textwidth}{!}{
\begin{tabular}{cccccccccc} 
\toprule
             & Cart.                    & Ball.                    & Quad.                  & Anymal                & Ant                  & Human.               & Franka                & Shadow                & Allegro                 \\ 
\midrule
PrefPPO-49   & {\textbf{499}{(0)}} & {\textbf{499}{(0)}} & -1.066(0.16)           & -1.861(0.03)          & 0.743(0.20)          & 0.457(0.09)          & 0.0044(0.00)          & 0.0746(0.02)          & 0.0125(0.003)           \\
PEBBLE-49  & \textbf{499}{(0)}      & \textbf{499}{(0)}         & -1.191(0.14)     & -1.3357(0.06)  & 5.9891(2.47)  & 3.67(1.32)    & 0.0453(0.01) & 0.2627(0.03) & 0.1467(0.03)   \\
SURF-49  & \textbf{499}{(0)}      & \textbf{499}{(0)}         & -1.202(0.03)    & -1.35(0.09)  & 0.874(0.18)  & 2.406(0.53)    & 0.0345(0.01) & 0.2338(0.03) & 0.2002(0.03)   \\
PrefPPO-150  & {\textbf{499}{(0)}} & {\textbf{499}{(0)}} & -0.959(0.15)           & -1.818(0.07)          & 0.171(0.05)          & 0.607(0.02)          & 0.0179(0.01)          & 0.0617(0.01)          & 0.0153(0.004)           \\
PEBBLE-150  & \textbf{499}{(0)}      & \textbf{499}{(0)}         & -1.059(0.07)     & -1.394(0.03)  & 7.257(2.34)  & 4.1417(1.11)   & 0.0532(0.02) & 0.269(0.02) & 0.2811(0.06)   \\
SURF-150  & \textbf{499}{(0)}      & \textbf{499}{(0)}         & -1.114(0.06)     & -1.383(0.03)  & 7.878(1.64)  & 4.312(1.18)    & 0.5285(0.18) & 0.2512(0.01) & 0.2727(0.05)   \\
PrefPPO-1.5k & {\textbf{499}{(0)}} & {\textbf{499}{(0)}} & -0.486(0.11)           & -1.417(0.21)          & 4.458(1.30)          & 1.329(0.33)          & 0.3248(0.12)          & 0.0488(0.01)          & 0.0284(0.005)           \\
PEBBLE-1.5k  & \textbf{499}{(0)}      & \textbf{499}{(0)}         & -0.529(0.14)     & -1.213(0.07)  & 9.364(0.71)  & 4.075(0.44)    & 0.1966(0.07) & 0.2538(0.03) & 0.2664(0.07)   \\
SURF-1.5k  & \textbf{499}{(0)}      & \textbf{499}{(0)}         & -0.308(0.06)     & -1.278(0.06) & 7.921(1.93)  & 3.577(0.24)    & 0.8032(0.27) & 0.2575(0.02) & 0.2283(0.05)   \\
PrefPPO-15k  & {\textbf{499}{(0)}} & {\textbf{499}{(0)}} & -0.250(0.06)           & -1.357(0.02)          & 4.626(0.57)          & 1.317(0.34)          & 0.0399(0.02)          & 0.0468(0.00)          & 0.0157(0.003)           \\
PEBBLE-15k  & \textbf{499}{(0)}      & \textbf{499}{(0)}         & -0.231(0.04)     & -0.73(0.21)  & 8.543(0.56)  & 6.162(0.97)    & 0.8613(0.16) & 0.246(0.02) & 0.2755(0.07)   \\
SURF-15k  & \textbf{499}{(0)}      & \textbf{499}{(0)}         & -0.266(0.02)     & -0.76(0.20)  & 7.859(1.45)  & 3.532(0.82)    & 0.5466(0.08) & 0.3199(0.05) & 0.2352(0.07)  \\
{\name}(Ours)       & {\textbf{499}{(0)}} & {\textbf{499}{(0)}} & {\textbf{-0.0195}(0.09)} & \textbf{-0.007}(0.35)          & {\textbf{12.04}(1.69)} & {\textbf{9.227}(0.93)} & {\textbf{0.9999}(0.24)} & {\textbf{13.231}(1.88)} & \textbf{25.030}(3.721)           \\
\hline
\hline
Eureka       & {499(0)}          & {499{(0)}} & -0.023(0.07)           & {-0.003(0.38)} & 10.86(0.85)          & 9.059(0.83)          & {0.9999(0.23)} & 11.532(1.38)          & {25.250(9.583)}  \\
\bottomrule
\end{tabular}
}
\caption{The final task score of all methods across different tasks in IssacGym. The values in parentheses represent the standard deviation.}
\label{tab: app-proxy}
\end{table}

\begin{table}
\centering
\resizebox{1.0\textwidth}{!}{
\begin{tabular}{cccccccccc} 
\toprule
         & Cart.  & Ball.  & Quad.         & Anymal       & Ant         & Human.      & Franka       & Shadow       & Allegro        \\ 
\midrule
{\name} w/o RT   & 499(0) & 499(0) & -0.0340(0.05) & -0.387(0.26) & 10.50(0.45) & 8.337(0.60) & 0.9999(0.25) & 10.769(2.30) & 25.641(9.46)   \\
{\name} w/o RTD  & 499(0) & 499(0) & -0.0216(0.14) & -0.009(0.38) & 10.53(0.39) & 9.419(2.10) & 1.0000(0.18) & 11.633(1.25) & 23.744(8.80)   \\
{\name} w/o RTDB & 499(0) & 499(0) & -0.0136(0.03) & -0.014(0.42) & 11.97(0.71) & 8.214(2.88) & 0.5129(0.06) & 13.663(1.83) & 25.386(3.42)   \\
OpenLoop & 499(0) & 499(0) & -0.0410(0.32) & -0.016(0.50) & 9.350(2.34) & 8.306(1.63) & 0.9999(0.22) & 9.476(2.44)  & 23.876(7.91)   \\
{\name}(Ours)   & 499(0) & 499(0) & -0.0195(0.09) & -0.007(0.35) & 12.04(1.69) & 9.227(0.93) & 0.9999(0.24) & 13.231(1.88) & 25.030(3.721)  \\
\bottomrule
\end{tabular}
}
\caption{Ablation studies on {\name} modules. The values in parentheses represent the standard deviation.}
\label{tab: app-ab}
\end{table}

\begin{table}
\centering
\resizebox{1.0\textwidth}{!}{
\begin{tabular}{cccccccccc} 
\toprule
             & Cart.  & Ball.  & Quad.         & Anymal       & Ant         & Human.      & Franka       & Shadow       & Allegro        \\ 
\midrule
PrefPPO-49   & 499(0) & 499(0) & -1.288(0.04)  & -1.833(0.05) & 0.281(0.06) & 0.855(0.24) & 0.0009(0.00) & 0.1178(0.03) & 0.1000(0.024)  \\
PrefPPO-150  & 499(0) & 499(0) & -1.288(0.02)  & -1.814(0.07) & 0.545(0.16) & 0.546(0.09) & 0.0012(0.00) & 0.0517(0.01) & 0.0544(0.010)  \\
PrefPPO-1.5k & 499(0) & 499(0) & -1.292(0.05)  & -1.583(0.13) & 2.235(0.63) & 2.480(0.59) & 0.0077(0.00) & 0.0495(0.01) & 0.0667(0.017)  \\
PrefPPO-15k  & 499(0) & 499(0) & -1.322(0.04)  & -1.611(0.12) & 3.694(0.86) & 1.867(0.19) & 0.0066(0.00) & 0.0543(0.01) & 0.1002(0.030)  \\
% Eureka       & 499(0) & 499(0) & -0.023(0.07)  & -0.003(0.38) & 10.86(0.85) & 9.059(0.83) & 0.9999(0.23) & 11.532(1.38) & 25.250(9.583)  \\
ICPL(Ours)       & 499(0) & 499(0) & -0.0195(0.09) & -0.007(0.35) & 12.04(1.69) & 9.227(0.93) & 0.9999(0.24) & 13.231(1.88) & 25.030(3.721)  \\
\bottomrule
\end{tabular}
}
\caption{The final task score of all methods across different tasks in IssacGym, where PrefPPO uses sparse rewards as the preference metric for the simulated teacher. The values in parentheses represent the standard deviation.}
\label{tab: app-sparse}
\end{table}

\section{Proxy Human Preference}

\subsection{Additional Results}
\label{app: add}
Due to the high variance in LLMs performance, we report the standard deviation across 5 experiments as a supplement, which is presented in Table \ref{tab: app-proxy} and Table \ref{tab: app-ab}. We also report the final task score of PrefPPO using sparse rewards as the preference metric for the simulated teacher in Table \ref{tab: app-sparse}. \rebuttal{Since ICPL involves new RL training in each iteration, it could be computationally expensive, we further provide the total training time (in hours) for all methods in Table~\ref{tab:total_time_horizontal} on the most computationally expensive task, Humanoid, which serves as a representative benchmark. As shown, although ICPL involves iterative reward generation and retraining, its computational cost is comparable to PEBBLE-1500, yet it achieves the best performance. Moreover, the entire ICPL pipeline completes within one day, making it a practical choice considering the performance gains.
}

\begin{table}[h]
\centering
\begin{tabular}{lccccccc}
\toprule
\textbf{Method} & PrefPPO-49 & PrefPPO-1500 & PEBBLE-49 & PEBBLE-1500 & Surf-49 & Surf-1500 & ICPL \\
\midrule
\textbf{Total Time (hrs)} & 1.9 & 2.2 & 15.5 & 16.6 & 10.3 & 12.5 & 16.4 \\
\bottomrule
\end{tabular}
\caption{\rebuttal{Total training time (in hours) for different methods.}}
\label{tab:total_time_horizontal}
\end{table}

We use a trial of the \textit{Humanoid} task to illustrate how {\name} progressively generated improved reward functions over successive iterations. The task description is ``to make the humanoid run as fast as possible''. Throughout five iterations, adjustments were made to the penalty terms and reward weightings. In the first iteration, the total reward was calculated as $0.5 \times \text{speed\_reward} + 0.25 \times \text{deviation\_reward} + 0.25 \times \text{action\_reward}$, yielding an RTS of 5.803. The speed reward and deviation reward motivate the humanoid to run fast, while the action reward promotes smoother motion.
In the second iteration, the weight of the speed reward was increased to 0.6, while the weights for deviation and action rewards were adjusted to 0.2 each, improving the RTS to 6.113. 
In the third iteration, the action penalty was raised and the reward weights were further modified to $0.7 \times \text{speed\_reward}$, $0.15 \times \text{deviation\_reward}$, and $0.15 \times \text{action\_reward}$, resulting in an RTS of 7.915.
During the fourth iteration, the deviation penalty was reduced to 0.35 and the action penalty was lowered, with the reward weights set to 0.8, 0.1, and 0.1 for speed, deviation, and action rewards, respectively. This change led to an RTS of 8.125.
Finally, in the fifth iteration, an additional upright reward term was incorporated, with the total reward calculated as $0.7 \times \text{speed\_reward} + 0.1 \times \text{deviation\_reward} + 0.1 \times \text{action\_reward} + 0.1 \times \text{upright\_reward}$. This adjustment produced the highest RTS of 8.232, allowing {\name} to generate reward functions that were more effectively aligned with the task description. Below are the specific reward functions produced at each iteration during one experiment. 

% First compute_reward function
\begin{tcolorbox}[breakable, colback=gray!5!white, colframe=gray!75!black, title=Humanoid Task: Reward Function with highest RTS (5.803) of Iteration 1]
\begin{lstlisting}[language=Python]
def compute_reward(root_states: torch.Tensor, actions: torch.Tensor) -> Tuple[torch.Tensor, Dict[str, torch.Tensor]]:
    velocity = root_states[: , 7:10]
    forward_velocity = velocity[:, 0]
    target_velocity = 5.0
    deviation_penalty = 0.5
    action_penalty = 0.1

    # Measure how fast the humanoid is going
    speed_reward = torch.exp((forward_velocity - target_velocity))

    # Penalize deviation from the target velocity
    deviation = torch.abs(forward_velocity - target_velocity)
    deviation_reward = torch.exp(-deviation_penalty * deviation)

    # Penalize actions, to induce smoother motion
    sum_actions = torch.sum(actions ** 2, dim=1)
    action_reward = torch.exp(-action_penalty * sum_actions)

    rewards = {'speed_reward': speed_reward, 'deviation_reward': deviation_reward, 'action_reward': action_reward }

    total_reward = 0.5 * speed_reward + 0.25 * deviation_reward + 0.25 * action_reward
    total_reward = total_reward.to(deviation.device)

    return total_reward, rewards
\end{lstlisting}
\end{tcolorbox}

% Second compute_reward function with highlighted differences
\begin{tcolorbox}[breakable, colback=gray!5!white, colframe=gray!75!black, title= Humanoid Task: Reward Function with highest RTS (6.113) of Iteration 2]
\begin{lstlisting}[language=Python]
def compute_reward(root_states: torch.Tensor, actions: torch.Tensor) -> Tuple[torch.Tensor, Dict[str, torch.Tensor]]:
    velocity = root_states[: , 7:10]
    forward_velocity = velocity[:, 0]
    target_velocity = 5.0
    deviation_penalty = 0.5
    action_penalty = 0.1

    # Measure how fast the humanoid is going
    speed_reward = torch.exp((forward_velocity - target_velocity))
    # Penalize deviation from the target velocity
    deviation = torch.abs(forward_velocity - target_velocity)
    deviation_reward = torch.exp(-deviation_penalty * deviation)
    # Penalize actions, to induce smoother motion
    sum_actions = torch.sum(actions ** 2, dim=1)
    action_reward = torch.exp(-action_penalty * sum_actions)

    rewards = {'speed_reward': speed_reward, 'deviation_reward': deviation_reward, 'action_reward': action_reward }

    # Adjusted weights in total reward calculation
    total_reward = 0.6 * speed_reward + 0.2 * deviation_reward + 0.2 * action_reward  # Changed weights
    total_reward = total_reward.to(deviation.device)

    return total_reward, rewards
\end{lstlisting}
\end{tcolorbox}

% Third compute_reward function with highlighted differences
\begin{tcolorbox}[breakable, colback=gray!5!white, colframe=gray!75!black, title=Humanoid Task: Reward Function with highest RTS (7.915) of Iteration 3]
\begin{lstlisting}[language=Python]
def compute_reward(root_states: torch.Tensor, actions: torch.Tensor) -> Tuple[torch.Tensor, Dict[str, torch.Tensor]]:
    velocity = root_states[: , 7:10]
    forward_velocity = velocity[:, 0]
    target_velocity = 5.0
    deviation_penalty = 0.5
    action_penalty = 0.15  # Increased action penalty

    # Measure how fast the humanoid is going
    speed_reward = torch.exp((forward_velocity - target_velocity))
    # Penalize deviation from the target velocity
    deviation = torch.abs(forward_velocity - target_velocity)
    deviation_reward = torch.exp(-deviation_penalty * deviation)
    # Penalize actions, to induce smoother motion
    sum_actions = torch.sum(actions ** 2, dim=1)
    action_reward = torch.exp(-action_penalty * sum_actions)

    rewards = {'speed_reward': speed_reward, 'deviation_reward': deviation_reward, 'action_reward': action_reward }

    # Adjusted weights in total reward calculation
    total_reward = 0.7 * speed_reward + 0.15 * deviation_reward + 0.15 * action_reward  # Changed weights
    total_reward = total_reward.to(deviation.device)

    return total_reward, rewards
\end{lstlisting}
\end{tcolorbox}

% Fourth compute_reward function with new upright penalty
\begin{tcolorbox}[breakable, colback=gray!5!white, colframe=gray!75!black, title=Humanoid Task: Reward Function with highest RTS (8.125) of Iteration 4]
\begin{lstlisting}[language=Python]
def compute_reward(root_states: torch.Tensor, actions: torch.Tensor) -> Tuple[torch.Tensor, Dict[str, torch.Tensor]]:
    velocity = root_states[:, 7:10]
    forward_velocity = velocity[:, 0]
    target_velocity = 5.0
    deviation_penalty = 0.35  # decrease penalty for deviation
    action_penalty = 0.1  # decrease penalty for action

    # Measure how fast the humanoid is going
    speed_reward = torch.exp((forward_velocity - target_velocity))
    # Penalize deviation from the target velocity
    deviation = torch.abs(forward_velocity - target_velocity)
    deviation_reward = torch.exp(-deviation_penalty * deviation)
    # Penalize actions, to induce smoother motion
    sum_actions = torch.sum(actions ** 2, dim=1)
    action_reward = torch.exp(-action_penalty * sum_actions)

    rewards = {'speed_reward': speed_reward, 'deviation_reward': deviation_reward, 'action_reward': action_reward }

    # Adjust the weights in total reward calculation
    total_reward = 0.8 * speed_reward + 0.1 * deviation_reward + 0.1 * action_reward # Changed weights
    total_reward = total_reward.to(deviation.device)

    return total_reward, rewards
\end{lstlisting}
\end{tcolorbox}

% Fourth compute_reward function with new upright penalty
\begin{tcolorbox}[breakable, colback=gray!5!white, colframe=gray!75!black, title=Humanoid Task: Reward Function with highest RTS (8.233) of Iteration 5]
\begin{lstlisting}[language=Python]
def compute_reward(root_states: torch.Tensor, actions: torch.Tensor, up_vec: torch.Tensor) -> Tuple[torch.Tensor, Dict[str, torch.Tensor]]:
    velocity = root_states[: , 7:10]
    forward_velocity = velocity[:, 0]
    target_velocity = 5.0
    deviation_penalty = 0.35  # Reduced deviation penalty
    action_penalty = 0.1  # Decreased action penalty

    # Measure how fast the humanoid is going
    speed_reward = torch.exp((forward_velocity - target_velocity))
    # Penalize deviation from the target velocity
    deviation = torch.abs(forward_velocity - target_velocity)
    deviation_reward = torch.exp(-deviation_penalty * deviation)
    # Penalize actions, to induce smoother motion
    sum_actions = torch.sum(actions ** 2, dim=1)
    action_reward = torch.exp(-action_penalty * sum_actions)
    # Reward for maintaining an upright position
    upright_penalty = 1.0  # New upright penalty for the humanoid
    upright_reward = torch.exp(-upright_penalty * (1 - up_vec[:, 2]))  # Added upright reward

    rewards = {'speed_reward': speed_reward, 'deviation_reward': deviation_reward, 'action_reward': action_reward, 'upright_reward': upright_reward }

    # Adjusted weights in total reward calculation
    total_reward = 0.7 * speed_reward + 0.1 * deviation_reward + 0.1 * action_reward + 0.1 * upright_reward  # Added upright reward to total
    total_reward = total_reward.to(deviation.device)

    return total_reward, rewards
\end{lstlisting}
\end{tcolorbox}

\section{Human-in-the-loop Preference}

\subsection{Recruitment Protocol}
\rebuttal{Participants were recruited through posters within the campus. Prior to participation, all volunteers were provided with an Information Sheet that clearly outlined: the purpose of the study, the tasks they would be asked to perform, the expected duration,
their right to withdraw at any time, how their data would be used and stored, and the compensation they would receive.
Only participants who gave informed consent in writing were included in the study. No personal identifiable information was collected. All data was anonymized and used exclusively for academic research purposes.}

\subsection{Demographic Data}
The participants in the human-in-the-loop preference experiments consisted of 7 individuals aged 19 to 30, including 2 women and 5 men. Their educational backgrounds included 2 undergraduate students and 5 graduate students. The 20 volunteers recruited to evaluate the performance of different methods were aged 23 to 28, comprising 5 women and 15 men, with 3 undergraduates and 17 graduate students.

\subsection{Human experiment setup}
\label{app: setup}
In ICPL experiments, each volunteer was assigned an account with a pre-configured environment to ensure smooth operation. After starting the experiment, LLMs generated the first iteration of reward functions. Once the reinforcement learning training was completed, videos corresponding to the policies derived from each reward function were automatically rendered. Volunteers compared the behaviors in the videos with the task descriptions and selected both the best and the worst-performing videos. They then entered the respective identifiers of these videos into the interactive interface and pressed ``Enter'' to proceed. The human preference was processed as an LLM prompt for generating feedback, leading to the next iteration of reward function generation.

This training-rendering-selection process was repeated across several iterations. At the end of the final iteration, the volunteers were asked to select the best video from those previously marked as good, designating it as the final result of the experiment. For IsaacGym tasks, the corresponding RTS was recorded as TS. It is important to note that, unlike proxy human preference experiments where the TS is the maximum RTS across iterations, in the human-in-the-loop preference experiment, TS refers to the highest RTS chosen by the human, as human selections are not always based on the maximum RTS at each iteration. Given that {\name} required reinforcement learning training in every iteration, each experiment lasted two to three days. Each volunteer was assigned a specific task and conducted five experiments, one for each task, with the highest TS being recorded as FTS in IsaacGym tasks.

\subsection{IsaacGym Tasks}
\label{sec: app-human}

We evaluate human-in-the-loop preference experiments on tasks in IsaacGym, including \textit{Quadcopter, Humanoid, Ant, ShadowHand, and AllegroHand}. In these experiments, volunteers were limited to comparing reward functions based solely on videos showcasing the final policies derived from each reward function.

In the \textit{Quadcopter} task, humans evaluate performance by observing whether the quadcopter moves quickly and efficiently, and whether it stabilizes in the final position. For the \textit{Humanoid} and \textit{Ant} tasks, where the task description is "make the ant/humanoid run as fast as possible," humans estimate speed by comparing the time taken to cover the same distance and assessing the movement posture. However, due to the variability in movement postures and directions, estimating speed can introduce inaccuracies. In the \textit{ShadowHand} and \textit{AllegroHand} tasks, where the goal is ``to make the hand spin the object to a target orientation,'' Humans find it challenging to calculate the precise difference between the current orientation and the target orientation at every moment, even though the target orientation is displayed nearby. Nevertheless, humans still can estimate the duration of effective rotations with the target orientation in the video, thus evaluating the performance of a single spin. Since the target orientation regenerates upon being reached, the frequency of target orientation changes can also aid in facilitating the assessment of evaluating performance.

Due to the lack of precise environmental data, volunteers cannot make absolutely accurate judgments during the experiments. For instance, in the \textit{Humanoid} task, robots may move in varying directions, which can introduce biases in volunteers' assessments of speed. However, volunteers are still able to filter out extremely poor results and select videos with relatively better performance. In most cases, the selected results closely align with those derived from proxy human preferences, enabling effective improvements in task performance.

Below is a specific case from the \textit{Humanoid} task that illustrates the potential errors humans may make during evaluation and the learning process of the reward function under this assumption. The reward task scores (RTS) chosen by the volunteer across five iterations are $4.521, 6.069, 6.814, 6.363, 6.983$. 

In the first iteration, the ground-truth task scores of each policy were $0.593, 2.744, 4.520, 0.192, 2.517, 5.937$, although the volunteer was unaware of these scores. Initially, the volunteer eliminated policies 0 and 3, as the robots in those videos primarily exhibited spinning behavior. Subsequently, the volunteer assessed the speed of the remaining robots based on how quickly a specific robot moved out of the field. The volunteer correctly identified that the robots in policies 1 and 4 were slightly slower. However, due to minor differences in the movement directions of the robots in policies 2 and 5, the volunteer mistakenly selected policy 2 as the best option, incorrectly concluding that the robot in policy 2 was faster.

Thus, the reward function selected in iteration 1 consists of several key components: velocity reward, upright reward, force penalty, unnatural pose penalty, and action penalty. These components not only promote faster training, which is the primary objective, but also encourage the maintenance of an upright pose. Additionally, the function penalizes excessive force usage, extreme joint angles, and large action values to foster smoother and more controlled movements.

In subsequent iterations, the volunteer effectively identified reward functions that exhibited relatively better and worse performance outcomes. Adjustments were made to the weights of each component, and specific temperature values were introduced for each. These modifications resulted in a more balanced reward structure, ensuring that critical aspects exert a stronger influence, thereby allowing for greater control over the learning dynamics and improving the agent's performance in achieving the task. Even in Iteration 4, the volunteer did not select the reward function with the highest RTS (6.813) but instead opted for the second-highest reward function (RTS = 6.363). Nevertheless, the reward function exhibited consistent improvement during these iterations.

Here we show the full reward function during the process. 

% First compute_reward function
\begin{tcolorbox}[breakable, colback=gray!5!white, colframe=gray!75!black, title=Humanoid Task: Reward Function chosen by volunteer with RTS (4.521) of Iteration 1]
\begin{lstlisting}[language=Python]
def compute_reward(
    velocity: torch.Tensor,
    dof_pos: torch.Tensor,
    dof_force_tensor: torch.Tensor,
    actions: torch.Tensor,
    up_vec: torch.Tensor,
    heading_vec: torch.Tensor
) -> Tuple[torch.Tensor, Dict[str, torch.Tensor]]:
    # Reward for forward velocity (run as fast as possible)
    velocity_reward = velocity[:, 0]
    
    # Encouragement for upright pose (penalize for deviation from vertical)
    upright_reward = up_vec[:, 2]
    
    # Penalize high force usage (energy efficiency)
    force_penalty = torch.sum(torch.abs(dof_force_tensor), dim=1)
    
    # Penalize unnatural joint positions (for instance, avoid extreme angles)
    unnatural_pose_penalty = torch.sum(torch.abs(dof_pos), dim=1)
    
    # Penalize for large actions (to promote smoother movement)
    action_penalty = torch.sum(torch.abs(actions), dim=1)
    
    # Normalize the rewards and penalties
    velocity_reward = torch.exp(velocity_reward) - 1
    upright_reward = torch.exp(upright_reward) - 1
    temperature = 1.0
    force_penalty = torch.exp(-force_penalty / temperature)
    unnatural_pose_penalty = torch.exp(-unnatural_pose_penalty / temperature)
    action_penalty = torch.exp(-action_penalty / temperature)
    
    # Combine the rewards and penalties into a single reward
    total_reward = (
        velocity_reward +
        0.5 * upright_reward -
        0.01 * force_penalty -
        0.01 * unnatural_pose_penalty -
        0.01 * action_penalty
    )
    
    # Return the total reward and each component for analysis
    reward_components = {
        "velocity_reward": velocity_reward,
        "upright_reward": upright_reward,
        "force_penalty": force_penalty,
        "unnatural_pose_penalty": unnatural_pose_penalty,
        "action_penalty": action_penalty
    }
    
    return total_reward, reward_components

\end{lstlisting}
\end{tcolorbox}

% Second compute_reward function with highlighted differences
\begin{tcolorbox}[breakable, colback=gray!5!white, colframe=gray!75!black, title= Humanoid Task: Reward Function chosen by volunteer with RTS (6.069) of Iteration 2]
\begin{lstlisting}[language=Python]
def compute_reward(
    velocity: torch.Tensor,
    dof_pos: torch.Tensor,
    dof_force_tensor: torch.Tensor,
    actions: torch.Tensor,
    up_vec: torch.Tensor,
    heading_vec: torch.Tensor
) -> Tuple[torch.Tensor, Dict[str, torch.Tensor]]:
    # Reward for forward velocity (run as fast as possible)
    velocity_reward = velocity[:, 0]
    velocity_temperature = 1.2  # increased slightly
    velocity_reward = torch.exp(velocity_reward / velocity_temperature) - 1
    
    # Encouragement for upright pose (penalize for deviation from vertical)
    upright_reward = up_vec[:, 2]
    upright_temperature = 0.5  # introduce a specific temperature
    upright_reward = torch.exp(upright_reward / upright_temperature) - 1
    
    # Penalize high force usage (energy efficiency)
    force_penalty = torch.sum(torch.abs(dof_force_tensor), dim=1)
    force_temperature = 0.1  # decreased to make it more significant
    force_penalty = torch.exp(-force_penalty / force_temperature)
    
    # Penalize unnatural joint positions (for instance, avoid extreme angles)
    unnatural_pose_penalty = torch.sum(torch.abs(dof_pos), dim=1)
    pose_temperature = 0.1  # decreased to make it more significant
    unnatural_pose_penalty = torch.exp(-unnatural_pose_penalty / pose_temperature)
    
    # Penalize for large actions (to promote smoother movement)
    action_penalty = torch.sum(torch.abs(actions), dim=1)
    action_temperature = 0.1  # decreased to make it more significant
    action_penalty = torch.exp(-action_penalty / action_temperature)
    
    # Combine the rewards and penalties into a single reward
    total_reward = (
        velocity_reward +
        0.5 * upright_reward -
        0.01 * force_penalty -
        0.01 * unnatural_pose_penalty -
        0.01 * action_penalty
    )
    
    # Return the total reward and each component for analysis
    reward_components = {
        "velocity_reward": velocity_reward,
        "upright_reward": upright_reward,
        "force_penalty": force_penalty,
        "unnatural_pose_penalty": unnatural_pose_penalty,
        "action_penalty": action_penalty
    }
    
    return total_reward, reward_components

\end{lstlisting}
\end{tcolorbox}

% Third compute_reward function with highlighted differences
\begin{tcolorbox}[breakable, colback=gray!5!white, colframe=gray!75!black, title= Humanoid Task: Reward Function chosen by volunteer with RTS (6.814) of Iteration 3]
\begin{lstlisting}[language=Python]
def compute_reward(
    velocity: torch.Tensor,
    dof_pos: torch.Tensor,
    dof_force_tensor: torch.Tensor,
    actions: torch.Tensor,
    up_vec: torch.Tensor,
    heading_vec: torch.Tensor
) -> Tuple[torch.Tensor, Dict[str, torch.Tensor]]:
    # Reward for forward velocity (run as fast as possible)
    velocity_reward = velocity[:, 0]
    velocity_temperature = 1.1  # minor adjustment
    velocity_reward = torch.exp(velocity_reward / velocity_temperature) - 1
    
    # Encouragement for upright pose (penalize for deviation from vertical)
    upright_reward = up_vec[:, 2]
    upright_temperature = 0.6  # slight adjustment
    upright_reward = torch.exp(upright_reward / upright_temperature) - 1
    
    # Penalize high force usage (energy efficiency)
    force_penalty = torch.sum(torch.abs(dof_force_tensor), dim=1)
    force_temperature = 0.15  # increased to try to make it effective
    force_penalty = torch.exp(-force_penalty / force_temperature)
    
    # Penalize unnatural joint positions (for instance, avoid extreme angles)
    unnatural_pose_penalty = torch.sum(torch.abs(dof_pos), dim=1)
    pose_temperature = 0.2  # increased to try to make it effective
    unnatural_pose_penalty = torch.exp(-unnatural_pose_penalty / pose_temperature)
    
    # Penalize for large actions (to promote smoother movement)
    action_penalty = torch.sum(torch.abs(actions), dim=1)
    action_temperature = 0.2  # increased to try to make it effective
    action_penalty = torch.exp(-action_penalty / action_temperature)
    
    # Combine the rewards and penalties into a single reward
    total_reward = (
        velocity_reward +
        0.5 * upright_reward -
        0.02 * force_penalty -  # increased slightly for more impact
        0.02 * unnatural_pose_penalty -  # increased slightly for more impact
        0.02 * action_penalty  # increased slightly for more impact
    )
    
    # Return the total reward and each component for analysis
    reward_components = {
        "velocity_reward": velocity_reward,
        "upright_reward": upright_reward,
        "force_penalty": force_penalty,
        "unnatural_pose_penalty": unnatural_pose_penalty,
        "action_penalty": action_penalty
    }
    
    return total_reward, reward_components

\end{lstlisting}
\end{tcolorbox}

% Fourth compute_reward function with highlighted differences
\begin{tcolorbox}[breakable, colback=gray!5!white, colframe=gray!75!black, title= Humanoid Task: Reward Function chosen by volunteer with RTS (6.363) of Iteration 4]
\begin{lstlisting}[language=Python]
def compute_reward(
    velocity: torch.Tensor,
    dof_pos: torch.Tensor,
    dof_force_tensor: torch.Tensor,
    actions: torch.Tensor,
    up_vec: torch.Tensor,
    heading_vec: torch.Tensor
) -> Tuple[torch.Tensor, Dict[str, torch.Tensor]]:
    # Reward for forward velocity (run as fast as possible)
    velocity_reward = velocity[:, 0]
    velocity_temperature = 1.05  # slight adjustment to refine the impact
    velocity_reward = torch.exp(velocity_reward / velocity_temperature) - 1
    
    # Encouragement for upright pose (penalize for deviation from vertical)
    upright_reward = up_vec[:, 2]
    upright_temperature = 0.65  # slight loosening for more upright reward
    upright_reward = torch.exp(upright_reward / upright_temperature) - 1
    
    # Penalize high force usage (energy efficiency)
    force_penalty = torch.sum(torch.abs(dof_force_tensor), dim=1)
    force_temperature = 0.2  # increased to make it more significant
    force_penalty = torch.exp(-force_penalty / force_temperature)
    
    # Penalize unnatural joint positions (for instance, avoid extreme angles)
    unnatural_pose_penalty = torch.sum(torch.abs(dof_pos), dim=1)
    pose_temperature = 0.25  # slight increase to make this component effective
    unnatural_pose_penalty = torch.exp(-unnatural_pose_penalty / pose_temperature)
    
    # Penalize for large actions (to promote smoother movement)
    action_penalty = torch.sum(torch.abs(actions), dim=1)
    action_temperature = 0.25  # slightly adjusted for more prominent constraint
    action_penalty = torch.exp(-action_penalty / action_temperature)
    
    # Combine the rewards and penalties into a single reward
    total_reward = (
        velocity_reward +
        0.5 * upright_reward -
        0.015 * force_penalty -  # slight increase for more impact
        0.015 * unnatural_pose_penalty -  # slight increase for more impact
        0.015 * action_penalty  # slight increase for more impact
    )
    
    # Return the total reward and each component for analysis
    reward_components = {
        "velocity_reward": velocity_reward,
        "upright_reward": upright_reward,
        "force_penalty": force_penalty,
        "unnatural_pose_penalty": unnatural_pose_penalty,
        "action_penalty": action_penalty
    }
    
    return total_reward, reward_components

\end{lstlisting}
\end{tcolorbox}

% Fourth best compute_reward function with highlighted differences
\begin{tcolorbox}[breakable, colback=gray!5!white, colframe=gray!75!black, title= Humanoid Task: Reward Function with best RTS (6.813) of Iteration 4(not chosen by volunteer)]
\begin{lstlisting}[language=Python]
def compute_reward(
    velocity: torch.Tensor,
    dof_pos: torch.Tensor,
    dof_force_tensor: torch.Tensor,
    actions: torch.Tensor,
    up_vec: torch.Tensor,
    heading_vec: torch.Tensor
) -> Tuple[torch.Tensor, Dict[str, torch.Tensor]]:
    # Reward for forward velocity (run as fast as possible)
    velocity_reward = velocity[:, 0]
    velocity_temperature = 1.15
    velocity_reward = torch.exp(velocity_reward / velocity_temperature) - 1
    
    # Encouragement for upright pose (penalize for deviation from vertical)
    upright_reward = up_vec[:, 2]
    upright_temperature = 0.55
    upright_reward = torch.exp(upright_reward / upright_temperature) - 1
    
    # Penalize high force usage (energy efficiency)
    force_penalty = torch.sum(torch.abs(dof_force_tensor), dim=1)
    force_temperature = 0.12
    force_penalty = torch.exp(-force_penalty / force_temperature)
    
    # Penalize unnatural joint positions (for instance, avoid extreme angles)
    unnatural_pose_penalty = torch.sum(torch.abs(dof_pos), dim=1)
    pose_temperature = 0.18
    unnatural_pose_penalty = torch.exp(-unnatural_pose_penalty / pose_temperature)
    
    # Penalize for large actions (to promote smoother movement)
    action_penalty = torch.sum(torch.abs(actions), dim=1)
    action_temperature = 0.18
    action_penalty = torch.exp(-action_penalty / action_temperature)
    
    # Combine the rewards and penalties into a single reward
    total_reward = (
        velocity_reward +
        0.5 * upright_reward -
        0.02 * force_penalty -
        0.02 * unnatural_pose_penalty -
        0.02 * action_penalty
    )
    
    # Return the total reward and each component for analysis
    reward_components = {
        "velocity_reward": velocity_reward,
        "upright_reward": upright_reward,
        "force_penalty": force_penalty,
        "unnatural_pose_penalty": unnatural_pose_penalty,
        "action_penalty": action_penalty
    }
    
    return total_reward, reward_components


\end{lstlisting}
\end{tcolorbox}

% Fifth compute_reward function with highlighted differences
\begin{tcolorbox}[breakable, colback=gray!5!white, colframe=gray!75!black, title= Humanoid Task: Reward Function chosen by volunteer with RTS (6.983) of Iteration 5]
\begin{lstlisting}[language=Python]
def compute_reward(
    velocity: torch.Tensor,
    dof_pos: torch.Tensor,
    dof_force_tensor: torch.Tensor,
    actions: torch.Tensor,
    up_vec: torch.Tensor,
    heading_vec: torch.Tensor
) -> Tuple[torch.Tensor, Dict[str, torch.Tensor]]:
    # Adjusted parameters based on analysis
    velocity_temperature = 1.1
    upright_temperature = 0.65
    force_temperature = 0.25
    pose_temperature = 0.3
    action_temperature = 0.3

    # Reward for forward velocity (run as fast as possible)
    velocity_reward = velocity[:, 0]
    velocity_reward = torch.exp(velocity_reward / velocity_temperature) - 1
    
    # Encouragement for upright pose (penalize for deviation from vertical)
    upright_reward = up_vec[:, 2]
    upright_reward = torch.exp(upright_reward / upright_temperature) - 1
    
    # Penalize high force usage (energy efficiency)
    force_penalty = torch.sum(torch.abs(dof_force_tensor), dim=1)
    force_penalty = torch.exp(-force_penalty / force_temperature)
    
    # Penalize unnatural joint positions (for instance, avoid extreme angles)
    unnatural_pose_penalty = torch.sum(torch.abs(dof_pos), dim=1)
    unnatural_pose_penalty = torch.exp(-unnatural_pose_penalty / pose_temperature)
    
    # Penalize for large actions (to promote smoother movement)
    action_penalty = torch.sum(torch.abs(actions), dim=1)
    action_penalty = torch.exp(-action_penalty / action_temperature)
    
    # Combine the rewards and penalties into a single reward
    total_reward = (
        velocity_reward +
        0.5 * upright_reward -
        0.02 * force_penalty -
        0.02 * unnatural_pose_penalty -
        0.02 * action_penalty
    )
    
    # Return the total reward and each component for analysis
    reward_components = {
        "velocity_reward": velocity_reward,
        "upright_reward": upright_reward,
        "force_penalty": force_penalty,
        "unnatural_pose_penalty": unnatural_pose_penalty,
        "action_penalty": action_penalty
    }
    
    return total_reward, reward_components

\end{lstlisting}
\end{tcolorbox}

\subsection{HumanoidJump Task} 
\label{app: jump}
In our study, we introduced a novel task: \textit{HumanoidJump}, with the task description being ``to make humanoid jump like a real human.'' The prompt of environment context in this task is shown in Prompt \ref{prompt: jump}.
\begin{lstlisting}[basicstyle=\fontfamily{\ttdefault}\scriptsize, breaklines=true,caption={Prompts of Environment Context in \textit{HumanoidJump} Task}\label{prompt: jump}]
class HumanoidJump(VecTask):
    """Rest of the environment definition omitted."""
    def compute_observations(self):
        self.gym.refresh_dof_state_tensor(self.sim)
        self.gym.refresh_actor_root_state_tensor(self.sim)
        self.gym.refresh_force_sensor_tensor(self.sim)
        self.gym.refresh_dof_force_tensor(self.sim)

        self.obs_buf[:], self.torso_position[:], 
        self.prev_torso_position[:], self.velocity_world[:], 
        self.angular_velocity_world[:], self.velocity_local[:], 
        self.angular_velocity_local[:], self.up_vec[:], 
        self.heading_vec[:], self.right_leg_contact_force[:], 
        self.left_leg_contact_force[:] = \
            compute_humanoid_jump_observations(
            self.obs_buf, self.root_states, self.torso_position,
            self.inv_start_rot, self.dof_pos, self.dof_vel, 
            self.dof_force_tensor, self.dof_limits_lower, 
            self.dof_limits_upper, self.dof_vel_scale, 
            self.vec_sensor_tensor, self.actions, 
            self.dt, self.contact_force_scale, 
            self.angular_velocity_scale, 
            self.basis_vec0, self.basis_vec1)
    
    def compute_humanoid_jump_observations(obs_buf, root_states, torso_position, inv_start_rot, dof_pos, dof_vel, dof_force, dof_limits_lower, dof_limits_upper, dof_vel_scale, sensor_force_torques, actions, dt, contact_force_scale, angular_velocity_scale, basis_vec0, basis_vec1):
        # type: (Tensor, Tensor, Tensor, Tensor, Tensor, Tensor, Tensor, Tensor, Tensor, float, Tensor, Tensor, float, float, float, Tensor, Tensor) -> Tuple[Tensor, Tensor, Tensor, Tensor, Tensor, Tensor, Tensor, Tensor, Tensor, Tensor, Tensor]
    
        prev_torso_position_new = torso_position.clone()
    
        torso_position = root_states[:, 0:3]
        torso_rotation = root_states[:, 3:7]
        velocity_world = root_states[:, 7:10]
        angular_velocity_world = root_states[:, 10:13]
    
        torso_quat, up_proj, up_vec, heading_vec = compute_heading_and_up_vec(
            torso_rotation, inv_start_rot, basis_vec0, basis_vec1, 2)
    
        velocity_local, angular_velocity_local, roll, pitch, yaw = compute_rot_new(
            torso_quat, velocity_world, angular_velocity_world)
    
        roll = normalize_angle(roll).unsqueeze(-1)
        yaw = normalize_angle(yaw).unsqueeze(-1)
        dof_pos_scaled = unscale(dof_pos, dof_limits_lower, dof_limits_upper)
        scale_angular_velocity_local = angular_velocity_local * angular_velocity_scale
    
        obs = torch.cat((root_states[:, 0:3].view(-1, 3), velocity_local, 
                         scale_angular_velocity_local,
                         yaw, roll, up_proj.unsqueeze(-1),
                         dof_pos_scaled, dof_vel * dof_vel_scale, 
                         dof_force * contact_force_scale, 
                         sensor_force_torques.view(-1, 12) * contact_force_scale, 
                         actions), dim=-1)
    
        right_leg_contact_force = sensor_force_torques[:, 0:3]
        left_leg_contact_force = sensor_force_torques[:, 6:9]
        
        abdomen_y_pos = dof_pos[:, 0]
        abdomen_z_pos = dof_pos[:, 1]
        abdomen_x_pos = dof_pos[:, 2]
        right_hip_x_pos = dof_pos[:, 3]
        right_hip_z_pos = dof_pos[:, 4]
        right_hip_y_pos = dof_pos[:, 5]
        right_knee_pos = dof_pos[:, 6]
        right_ankle_x_pos = dof_pos[:, 7]
        right_ankle_y_pos = dof_pos[:, 8]
        left_hip_x_pos = dof_pos[:, 9]
        left_hip_z_pos = dof_pos[:, 10]
        left_hip_y_pos = dof_pos[:, 11]
        left_knee_pos = dof_pos[:, 12]
        left_ankle_x_pos = dof_pos[:, 13]
        left_ankle_y_pos = dof_pos[:, 14]
        right_shoulder1_pos = dof_pos[:, 15]
        right_shoulder2_pos = dof_pos[:, 16]
        right_elbow_pos = dof_pos[:, 17]
        left_shoulder1_pos = dof_pos[:, 18]
        left_shoulder2_pos = dof_pos[:, 19]
        left_elbow_pos = dof_pos[:, 20]
    
        right_shoulder1_action = actions[:, 15]
        right_shoulder2_action = actions[:, 16]
        right_elbow_action = actions[:, 17]
        left_shoulder1_action = actions[:, 18]
        left_shoulder2_action = actions[:, 19]
        left_elbow_action = actions[:, 20]
        
        return obs, torso_position, prev_torso_position_new, velocity_world, 
               angular_velocity_world, velocity_local, scale_angular_velocity_local, 
               up_vec, heading_vec, right_leg_contact_force, left_leg_contact_force
\end{lstlisting}

\subsubsection{Reward functions.} 
\label{app: reward}
We show the reward functions in a trial that successfully evolved a human-like jump: bending both legs to jump.
Initially, the reward function focused on encouraging vertical movement while penalizing horizontal displacement, high contact force usage, and improper joint movements. Over time, the scaling factors for the rewards and penalties were gradually adjusted by changing the temperature parameters in the exponential scaling. These adjustments aimed to enhance the model's sensitivity to different movement behaviors.
For example, the vertical movement reward's temperature was reduced, leading to more precise rewards for positive vertical movements. Similarly, the horizontal displacement penalty was fine-tuned by modifying its temperature across iterations, either decreasing or increasing the penalty's impact on lateral movements. The contact force penalty evolved by decreasing its temperature to penalize excessive force usage more strongly, especially in the later iterations, making the task more sensitive to leg contact forces. Finally, the joint usage reward was refined by adjusting the temperature to either encourage or discourage certain joint behaviors, with more focus on leg extension and contraction patterns.
Overall, the changes primarily revolved around adjusting the sensitivity of different components, refining the balance between rewards and penalties to better align the humanoid's behavior with the desired jumping performance.
% First compute_reward function
\begin{tcolorbox}[breakable, colback=gray!5!white, colframe=gray!75!black, title=HumanoidJump Task: Reward Function of Iteration 1]
\begin{lstlisting}[language=Python]
def compute_reward(torso_position: torch.Tensor, prev_torso_position: torch.Tensor, velocity_world: torch.Tensor, 
                   right_leg_contact_force: torch.Tensor, left_leg_contact_force: torch.Tensor, dof_pos: torch.Tensor) -> Tuple[torch.Tensor, Dict[str, torch.Tensor]]:
    # Ensure all tensors are on the same device
    device = torso_position.device

    # Compute vertical torso movement reward
    vertical_movement = torso_position[:, 2] - prev_torso_position[:, 2]
    vertical_movement_reward = torch.clamp(vertical_movement, min=0.0)  # Reward positive vertical movement
    vertical_movement_reward = torch.exp(vertical_movement_reward / 0.1)  # Use exponential scaling with temperature
    
    # Compute horizontal displacement penalty
    horizontal_displacement = torch.sum(torch.abs(torso_position[:, :2] - prev_torso_position[:, :2]), dim=-1)
    horizontal_displacement_penalty = torch.exp(-horizontal_displacement / 0.1)  # Penalize large movements with temperature

    # Compute leg forces usage reward
    contact_force_usage = torch.sum(torch.abs(right_leg_contact_force) + torch.abs(left_leg_contact_force), dim=-1)
    contact_force_usage_penalty = torch.exp(-contact_force_usage / 10.0)  # Penalize high contact force usage with temperature

    # Compute joint usage reward (encourages proper leg extension and contraction)
    leg_joints_indices = torch.tensor([6, 7, 8, 12, 13, 14], device=device)  # Indices of leg joints 
    leg_joint_usage = torch.mean(dof_pos[:, leg_joints_indices], dim=-1)
    leg_joint_usage_reward = torch.exp(-torch.abs(leg_joint_usage) / 0.1)  # Encourage movements from neutral position

    # Sum all rewards and penalties
    total_reward = vertical_movement_reward + horizontal_displacement_penalty + contact_force_usage_penalty + leg_joint_usage_reward

    # Create a dictionary for individual reward components
    reward_components = {
        'vertical_movement_reward': vertical_movement_reward,
        'horizontal_displacement_penalty': horizontal_displacement_penalty,
        'contact_force_usage_penalty': contact_force_usage_penalty,
        'leg_joint_usage_reward': leg_joint_usage_reward
    }
    
    return total_reward, reward_components
\end{lstlisting}
\end{tcolorbox}

% Second compute_reward function 
\begin{tcolorbox}[breakable, colback=gray!5!white, colframe=gray!75!black, title= HumanoidJump Task: Reward Function of Iteration 2]
\begin{lstlisting}[language=Python]
def compute_reward(
    torso_position: torch.Tensor,
    prev_torso_position: torch.Tensor,
    velocity_world: torch.Tensor,
    right_leg_contact_force: torch.Tensor,
    left_leg_contact_force: torch.Tensor,
    dof_pos: torch.Tensor
) -> Tuple[torch.Tensor, Dict[str, torch.Tensor]]:
    # Ensure all tensors are on the same device
    device = torso_position.device

    # Compute vertical torso movement reward
    vertical_movement = torso_position[:, 2] - prev_torso_position[:, 2]
    vertical_movement_reward = torch.clamp(vertical_movement, min=0.0)  # Reward positive vertical movement
    temperature_vertical = 0.05
    vertical_movement_reward = torch.exp(vertical_movement_reward / temperature_vertical)  # Use exponential scaling

    # Compute horizontal displacement penalty
    horizontal_displacement = torch.sum(torch.abs(torso_position[:, :2] - prev_torso_position[:, :2]), dim=-1)
    temperature_horizontal = 0.1
    horizontal_displacement_penalty = torch.exp(-horizontal_displacement / temperature_horizontal)  # Penalize large movements with temperature

    # Compute leg forces usage reward
    contact_force_usage = torch.sum(torch.abs(right_leg_contact_force) + torch.abs(left_leg_contact_force), dim=-1)
    temperature_force = 1.0
    contact_force_usage_penalty = torch.exp(-contact_force_usage / temperature_force)  # Penalize high contact force usage with temperature

    # Compute joint usage reward (encourages proper leg extension and contraction)
    leg_joints_indices = torch.tensor([6, 7, 8, 12, 13, 14], device=device)  # Indices of leg joints 
    leg_joint_usage = torch.mean(dof_pos[:, leg_joints_indices], dim=-1)
    temperature_joints = 0.05
    leg_joint_usage_reward = torch.exp(-torch.abs(leg_joint_usage) / temperature_joints)  # Encourage movements from the neutral position

    # Sum all rewards and penalties
    total_reward = vertical_movement_reward + horizontal_displacement_penalty + contact_force_usage_penalty + leg_joint_usage_reward

    # Create a dictionary for individual reward components
    reward_components = {
        'vertical_movement_reward': vertical_movement_reward,
        'horizontal_displacement_penalty': horizontal_displacement_penalty,
        'contact_force_usage_penalty': contact_force_usage_penalty,
        'leg_joint_usage_reward': leg_joint_usage_reward
    }

    return total_reward, reward_components
\end{lstlisting}
\end{tcolorbox}

% Third compute_reward function 
\begin{tcolorbox}[breakable, colback=gray!5!white, colframe=gray!75!black, title=HumanoidJump Task: Reward Function of Iteration 3]
\begin{lstlisting}[language=Python]
def compute_reward(
    torso_position: torch.Tensor,
    prev_torso_position: torch.Tensor,
    velocity_world: torch.Tensor,
    right_leg_contact_force: torch.Tensor,
    left_leg_contact_force: torch.Tensor,
    dof_pos: torch.Tensor
) -> Tuple[torch.Tensor, Dict[str, torch.Tensor]]:
    # Ensure all tensors are on the same device
    device = torso_position.device

    # Compute vertical torso movement reward
    vertical_movement = torso_position[:, 2] - prev_torso_position[:, 2]
    vertical_movement_reward = torch.clamp(vertical_movement, min=0.0)  # Reward positive vertical movement
    temperature_vertical = 0.05
    vertical_movement_reward = torch.exp(vertical_movement_reward / temperature_vertical)  # Use exponential scaling

    # Compute horizontal displacement penalty
    horizontal_displacement = torch.sum(torch.abs(torso_position[:, :2] - prev_torso_position[:, :2]), dim=-1)
    temperature_horizontal = 0.05  # Adjusted temperature for finer tuning
    horizontal_displacement_penalty = torch.exp(-horizontal_displacement / temperature_horizontal)  # Penalize large movements

    # Compute leg forces usage reward
    contact_force_usage = torch.sum(torch.abs(right_leg_contact_force) + torch.abs(left_leg_contact_force), dim=-1)
    temperature_force = 5.0  # Adjusted to make contact force usage more noticeable
    contact_force_usage_penalty = torch.exp(-contact_force_usage / temperature_force)  # Penalize high contact force usage

    # Compute joint usage reward (encourages proper leg extension and contraction)
    leg_joints_indices = torch.tensor([6, 7, 8, 12, 13, 14], device=device)  # Indices of leg joints 
    leg_joint_usage = torch.mean(dof_pos[:, leg_joints_indices], dim=-1)
    temperature_joints = 0.05
    leg_joint_usage_reward = torch.exp(-torch.abs(leg_joint_usage) / temperature_joints)  # Encourage movements from the neutral position

    # Sum all rewards and penalties
    total_reward = vertical_movement_reward + horizontal_displacement_penalty + contact_force_usage_penalty + leg_joint_usage_reward

    # Create a dictionary for individual reward components
    reward_components = {
        'vertical_movement_reward': vertical_movement_reward,
        'horizontal_displacement_penalty': horizontal_displacement_penalty,
        'contact_force_usage_penalty': contact_force_usage_penalty,
        'leg_joint_usage_reward': leg_joint_usage_reward
    }

    return total_reward, reward_components
\end{lstlisting}
\end{tcolorbox}

% Fourth compute_reward function 
\begin{tcolorbox}[breakable, colback=gray!5!white, colframe=gray!75!black, title=HumanoidJump Task: Reward Function of Iteration 4]
\begin{lstlisting}[language=Python]
def compute_reward(
    torso_position: torch.Tensor,
    prev_torso_position: torch.Tensor,
    velocity_world: torch.Tensor,
    right_leg_contact_force: torch.Tensor,
    left_leg_contact_force: torch.Tensor,
    dof_pos: torch.Tensor
) -> Tuple[torch.Tensor, Dict[str, torch.Tensor]]:
    # Ensure all tensors are on the same device
    device = torso_position.device

    # Compute vertical torso movement reward
    vertical_movement = torso_position[:, 2] - prev_torso_position[:, 2]
    vertical_movement_reward = torch.clamp(vertical_movement, min=0.0)  # Reward positive vertical movement
    temperature_vertical = 0.04  # Adjusted temperature for better sensitivity
    vertical_movement_reward = torch.exp(vertical_movement_reward / temperature_vertical)  # Use exponential scaling

    # Compute horizontal displacement penalty
    horizontal_displacement = torch.sum(torch.abs(torso_position[:, :2] - prev_torso_position[:, :2]), dim=-1)
    temperature_horizontal = 0.1  # Increased temperature to penalize horizontal movement more 
    horizontal_displacement_penalty = torch.exp(-horizontal_displacement / temperature_horizontal)  # Penalize large movements

    # Compute leg forces usage reward
    contact_force_usage = torch.sum(torch.abs(right_leg_contact_force) + torch.abs(left_leg_contact_force), dim=-1)
    temperature_force = 0.1  # Significantly increase sensitivity to contact forces
    contact_force_usage_penalty = torch.exp(-contact_force_usage / temperature_force)  # Penalize high contact force usage

    # Compute joint usage reward (encourages proper leg extension and contraction)
    leg_joints_indices = torch.tensor([6, 7, 8, 12, 13, 14], device=device)  # Indices of leg joints 
    leg_joint_usage = torch.mean(dof_pos[:, leg_joints_indices], dim=-1)
    temperature_joints = 0.02  # Adjusted for joint usage sensitivity
    leg_joint_usage_reward = torch.exp(-torch.abs(leg_joint_usage) / temperature_joints)  # Encourage movements from the neutral position

    # Sum all rewards and penalties
    total_reward = vertical_movement_reward + horizontal_displacement_penalty + contact_force_usage_penalty + leg_joint_usage_reward

    # Create a dictionary for individual reward components
    reward_components = {
        'vertical_movement_reward': vertical_movement_reward,
        'horizontal_displacement_penalty': horizontal_displacement_penalty,
        'contact_force_usage_penalty': contact_force_usage_penalty,
        'leg_joint_usage_reward': leg_joint_usage_reward
    }

    return total_reward, reward_components
\end{lstlisting}
\end{tcolorbox}

% Fifth compute_reward function 
\begin{tcolorbox}[breakable, colback=gray!5!white, colframe=gray!75!black, title=Humanoid Task: Reward Function of Iteration 5]
\begin{lstlisting}[language=Python]
def compute_reward(
    torso_position: torch.Tensor,
    prev_torso_position: torch.Tensor,
    velocity_world: torch.Tensor,
    right_leg_contact_force: torch.Tensor,
    left_leg_contact_force: torch.Tensor,
    dof_pos: torch.Tensor
) -> Tuple[torch.Tensor, Dict[str, torch.Tensor]]:
    # Ensure all tensors are on the same device
    device = torso_position.device

    # Compute vertical torso movement reward
    vertical_movement = torso_position[:, 2] - prev_torso_position[:, 2]
    vertical_movement_reward = torch.clamp(vertical_movement, min=0.0)  # Reward positive vertical movement
    temperature_vertical = 0.04  # Adjusted temperature for better sensitivity
    vertical_movement_reward = torch.exp(vertical_movement_reward / temperature_vertical)  # Use exponential scaling

    # Compute horizontal displacement penalty
    horizontal_displacement = torch.sum(torch.abs(torso_position[:, :2] - prev_torso_position[:, :2]), dim=-1)
    temperature_horizontal = 0.05  # Decreased temperature for more sensitivity
    horizontal_displacement_penalty = torch.exp(-horizontal_displacement / temperature_horizontal)  # Penalize large movements

    # Compute leg forces usage penalty (Rewritten to reduce contact force)
    contact_force_usage = torch.sum(torch.abs(right_leg_contact_force) + torch.abs(left_leg_contact_force), dim=-1)
    temperature_force = 0.5  # Adjusted to penalize contact force usage
    contact_force_usage_penalty = torch.exp(-contact_force_usage / temperature_force)  # Penalize high contact force usage

    # Compute joint usage reward (encourages proper leg extension and contraction)
    leg_joints_indices = torch.tensor([6, 7, 8, 12, 13, 14], device=device)  # Indices of leg joints 
    leg_joint_usage = torch.mean(torch.abs(dof_pos[:, leg_joints_indices]), dim=-1)
    temperature_joints = 0.02  # Adjusted for joint usage sensitivity
    leg_joint_usage_reward = torch.exp(-leg_joint_usage / temperature_joints)  # Encourage movements from the neutral position

    # Sum all rewards and penalties
    total_reward = vertical_movement_reward + horizontal_displacement_penalty + contact_force_usage_penalty + leg_joint_usage_reward

    # Create a dictionary for individual reward components
    reward_components = {
        'vertical_movement_reward': vertical_movement_reward,
        'horizontal_displacement_penalty': horizontal_displacement_penalty,
        'contact_force_usage_penalty': contact_force_usage_penalty,
        'leg_joint_usage_reward': leg_joint_usage_reward
    }

    return total_reward, reward_components
\end{lstlisting}
\end{tcolorbox}

% Sixth compute_reward function
\begin{tcolorbox}[breakable, colback=gray!5!white, colframe=gray!75!black, title=Humanoid Task: Reward Function of Iteration 6]
\begin{lstlisting}[language=Python]
def compute_reward(
    torso_position: torch.Tensor,
    prev_torso_position: torch.Tensor,
    velocity_world: torch.Tensor,
    right_leg_contact_force: torch.Tensor,
    left_leg_contact_force: torch.Tensor,
    dof_pos: torch.Tensor
) -> Tuple[torch.Tensor, Dict[str, torch.Tensor]]:
    # Ensure all tensors are on the same device
    device = torso_position.device

    # Compute vertical torso movement reward
    vertical_movement = torso_position[:, 2] - prev_torso_position[:, 2]
    vertical_movement_reward = torch.clamp(vertical_movement, min=0.0)  # Reward positive vertical movement
    temperature_vertical = 0.03  # Fine-tuned temperature for better sensitivity
    vertical_movement_reward = torch.exp(vertical_movement_reward / temperature_vertical)  # Use exponential scaling

    # Compute horizontal displacement penalty
    horizontal_displacement = torch.sum(torch.abs(torso_position[:, :2] - prev_torso_position[:, :2]), dim=-1)
    temperature_horizontal = 0.04  # Decreased temperature for more sensitivity
    horizontal_displacement_penalty = torch.exp(-horizontal_displacement / temperature_horizontal)  # Penalize large movements

    # Compute leg forces usage penalty (encourage minimal contact force)
    contact_force_usage = torch.sum(torch.abs(right_leg_contact_force) + torch.abs(left_leg_contact_force), dim=-1)
    temperature_force = 0.5  # Adjusted to penalize contact force usage
    contact_force_usage_penalty = torch.exp(-contact_force_usage / temperature_force)  # Penalize high contact force usage

    # Compute joint usage reward (encourages proper leg extension and contraction)
    leg_joints_indices = torch.tensor([6, 7, 8, 12, 13, 14], device=device)  # Indices of leg joints 
    leg_joint_usage = torch.mean(torch.abs(dof_pos[:, leg_joints_indices]), dim=-1)
    temperature_joints = 0.02  # Fine-tuned for joint usage sensitivity
    leg_joint_usage_reward = torch.exp(-torch.abs(leg_joint_usage) / temperature_joints)  # Encourage movements from the neutral position

    # Sum all rewards and penalties
    total_reward = vertical_movement_reward + horizontal_displacement_penalty + contact_force_usage_penalty + leg_joint_usage_reward

    # Create a dictionary for individual reward components
    reward_components = {
        'vertical_movement_reward': vertical_movement_reward,
        'horizontal_displacement_penalty': horizontal_displacement_penalty,
        'contact_force_usage_penalty': contact_force_usage_penalty,
        'leg_joint_usage_reward': leg_joint_usage_reward
    }

    return total_reward, reward_components
\end{lstlisting}
\end{tcolorbox}

\end{document}